\documentclass[lettersize,journal,twoside]{IEEEtran}
\usepackage{amsmath,amsfonts}
\usepackage{algorithmic}
\usepackage{algorithm}
\usepackage{array}
\usepackage[caption=false,font=normalsize,labelfont=sf,textfont=sf]{subfig}
\usepackage{textcomp}
\usepackage{stfloats}
\usepackage{url}
\usepackage{verbatim}
\usepackage{graphicx}
\usepackage{cite}
\usepackage{pifont}
\usepackage{amsmath}
\usepackage{multirow}
\usepackage{makecell}
\usepackage{amsthm}
\usepackage[dvipsnames]{xcolor}
\DeclareMathOperator*{\argmin}{arg\,min}

\begin{document}
\title{Recovering Pulse Waves from Video Using \\Deep Unrolling and Deep Equilibrium Models}

\author{Vineet R. Shenoy\thanks{Vineet R. Shenoy and Rama Chellappa are with the Johns Hopkins University, Baltimore, MD, USA (e-mail: vshenoy4@jhu.edu, rchella4@jhu.edu). VS Performed this work as an intern at MERL.}, Suhas Lohit, Hassan Mansour, Rama Chellappa, Tim K. Marks \thanks{Suhas Lohit, Hassan Mansour, and Tim K. Marks are with Mitsubishi Electric Research Laboratories (MERL), Cambridge, MA, USA (e-mail: \{slohit, mansour, tmarks\}@merl.com).}}

\markboth{}{Shenoy \MakeLowercase{et al.}: Recovering Pulse Waves from Video Using Deep Unrolling and Deep Equilibrium Models }

\newcommand\norm[1]{\left\lVert#1\right\rVert}
\newcommand\normx[1]{\left\Vert#1\right\Vert}

\newcommand{\cmark}{\ding{51}}
\newcommand{\xmark}{\ding{55}}
\newenvironment{psmallmatrix}
  {\left[\begin{smallmatrix}}
  {\end{smallmatrix}\right]}
\newtheorem{theorem}{Theorem}
\IEEEpubidadjcol
\maketitle
\begin{abstract}
Camera-based monitoring of vital signs, also known as imaging photoplethysmography (iPPG), has seen applications in driver-monitoring, perfusion assessment in surgical settings, affective computing, and more. iPPG involves sensing the  underlying cardiac pulse from video of the skin and estimating vital signs such as the heart rate or a full pulse waveform. Some previous iPPG methods impose model-based sparse priors on the pulse signals and use iterative optimization for pulse wave recovery, while others use end-to-end black-box deep learning methods. In contrast, we introduce methods that combine signal processing and deep learning methods in an inverse problem framework. Our methods estimate the underlying pulse signal and heart rate from facial video by learning deep-network-based denoising operators that leverage deep algorithm unfolding and deep equilibrium models. Experiments show that our methods can denoise an acquired signal from the face and infer the correct underlying pulse rate, achieving state-of-the-art heart rate estimation performance on well-known benchmarks, all with less than one-fifth the number of learnable parameters as the closest competing method. 
 
\end{abstract}

\section{Introduction}
\label{sec:intro}

\begin{figure*}[!ht]
    \centering
    \includegraphics[width=0.75\textwidth]{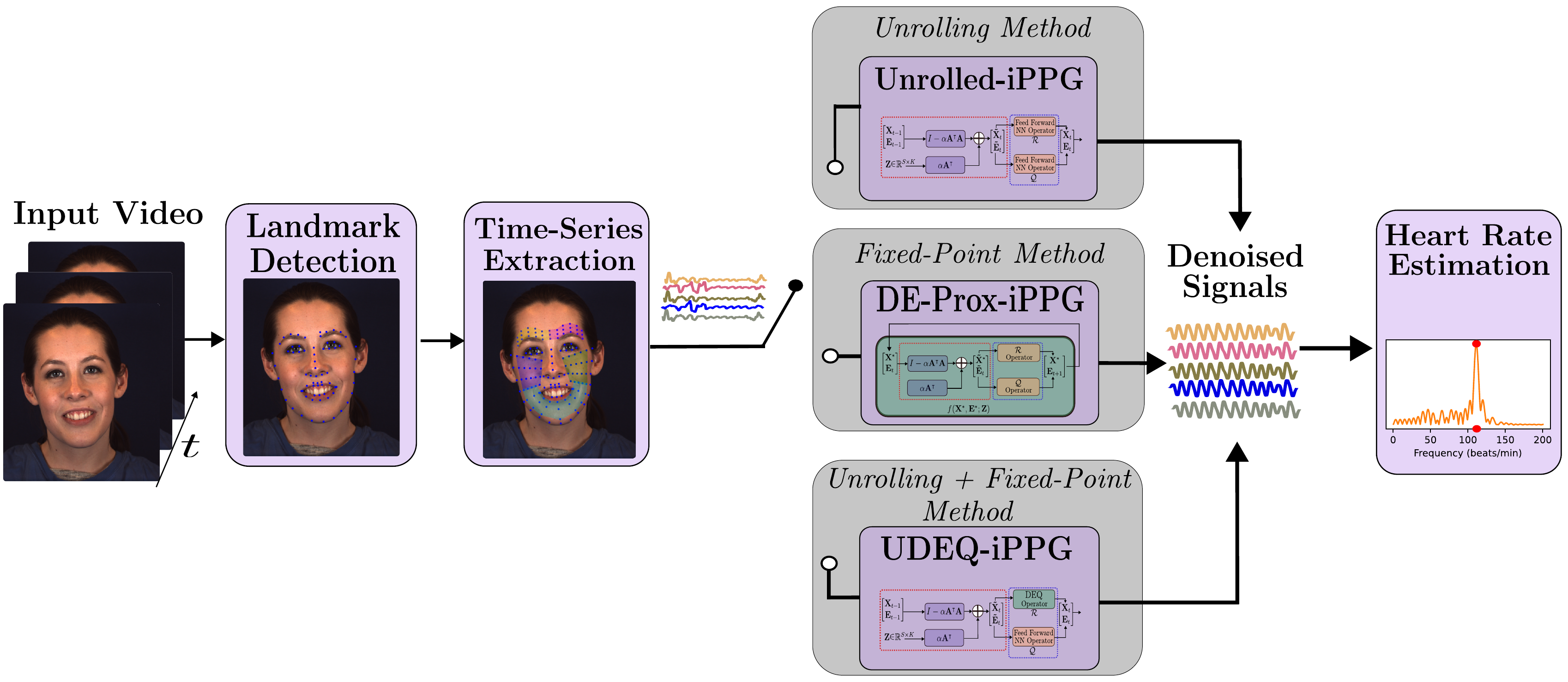}
    \caption{Our full pipeline. We pass the input video frames through a face and landmark detector, extrapolate additional facial landmarks, and obtain a time series based on pixel intensities from five segmented face regions. We then pass our raw signals to either Unrolled iPPG, DE-Prox-iPPG, or UDEQ-iPPG which separates the signal from the noise. Finally, we sum the power spectrum coefficients across individual bins for all five signals, and select the bin with the highest power as our heart-rate estimate.}
    \label{fig:hr_pipeline}
\end{figure*}
Physiological measurements using cameras~\cite{acmMcDuff23,liu2022rppg} have emerged as an exciting research topic, with many recent techniques focusing on imaging photoplethysmography (iPPG) for heart rate estimation. In this task, given a camera recording of the skin (e.g., the face), the pulse wave is reconstructed and subsequently heart rate is determined---all without contact with the skin.  The digital blood volume pulse signal, manifested as reflections of incident light on pulsatile blood flow, is weak in the presence of noise signals, and the main challenge is to separate the signal from the motion, lighting, and quantization noise. Pixel intensity changes due to the pulsatile signal are weak compared to those due to noise sources such as motion, lighting variations, and camera sensor quantization. The main challenge is to separate the signal from the noise caused by these and other factors. Developing robust solutions to iPPG waveform recovery could improve monitoring of sick patients~\cite{aarts2013non}, monitor drivers in automotive applications~\cite{autosparseppg}, and improve security in face recognition~\cite{liu2018remote}.

Early methods to solve this problem were modular and based on signal processing techniques; these methods cropped the face and defined one or more regions of interest, followed by a signal extraction technique. These include blind source separation methods \cite{poh2010non,lewandowska} and later improvements used chrominance information~\cite{chrom} and skin reflection properties~\cite{pbv}.  Another class of model-driven methods was based on sparse optimization~\cite{sparseppg,autosparseppg}; the authors assumed that the pulse signal was composed of a sparse set of frequency coefficients and that the noise was present in a sparse set of facial regions. The authors then solved for a pulse vector that is sparse in the frequency domain.  While these hand-crafted signal-processing methods set a strong foundation for future work, they could not learn signal representations from data to improve pulse rate estimation, especially under challenging recording scenarios.

Almost all recent data-driven iPPG methods are based on deep learning. After initial work~\cite{chen2018deepphys} modeled the nonlinear interactions between the pulse signal and noise, deep learning methods were further refined with temporal shift models~\cite{liu2020tscan}, spatio-temporal models~\cite{yu2019physnet, yu2022physformer}, and attention~\cite{nowara2021benefit}. Other deep neural network methods use meta-learning~\cite{liu2021metaphys, lee2020meta}, efficient computation~\cite{liu2023efficientphys}, and federated learning~\cite{liu2022federated}. With recent interest in learning without labels, unsupervised/self-supervised methods such as \cite{Gideon_2021_ICCV, sun2022contrast, self_supervised_rppg} have used contrastive learning and masked autoencoders to update network parameters without a supervisory signal. 

Compared to earlier iPPG methods, the deep-learning-based methods have a disadvantage in that they fundamentally lack interpretability. In this paper, we show that integrating both signal processing and learning-based techniques not only achieves strong iPPG performance, but also puts deep learning techniques on sound signal-processing foundations. 

To this end, our work expands on the optimization-based signal recovery algorithms while integrating lightweight deep learning components that play specialized roles for pulse signal recovery. Similar to~\cite{autosparseppg}, we model pulse signal recovery as an optimization problem that minimizes a data fidelity term and a prior-promoting penalty term, and solve the resulting optimization problem using ISTA-style~\cite{fista} forward-backward splitting algorithms. However, using hand-crafted priors and their corresponding proximal operators does not always capture the complex and nonlinear interactions between frequencies in the pulse signal or the noise that is present in the environment. Therefore, we impose \textit{implicit} priors on the heartbeat signal and the noise by training deep neural networks that function as denoising operators. Our approach therefore maintains the algorithmic structure of optimization-based methods while benefiting from data-driven  deep learning. 

We apply our insights to develop three different algorithms. First, we present \textbf{Unrolled iPPG}, which learns the denoising operators of the signal and noise priors by ``unrolling" proximal gradient descent~\cite{monga-unrolling} for a \textit{predetermined} number of iterations, applying a single forward pass through the structure-promoting operators of each of the signal and noise at each iteration. A potential drawback of Unrolled iPPG is that it applies only a single denoising step in each of the unrolled iterations, and the number of iterations must be decided in advance. This limitation can be overcome using fixed-point theory and deep equilibrium (DEQ) models~\cite{bai2019deep}, which use fixed-point solvers to effectively iterate the unrolling steps until the output converges. Incorporating these ideas, we develop \textbf{Deep Equilibrium-Proximal iPPG (DE-Prox-iPPG)}, which iterates \textit{both} the gradient update and learned proximal/denoising steps through a solver until a fixed-point is reached~\cite{gilton-inverse}. 

Our experiments show that this method, which makes fixed-point assumptions on both the signal and noise, has its own weaknesses. Given that the pulsatile signal is a realization of the quasi-periodic nature of the beating heart, we assume it is sufficiently structured to be represented as a fixed point of some operator. However, the noise signal, which includes noise due to numerous sources including motion and illumination variations,  may not admit a fixed-point representation. Consequently, we introduce \textbf{Unrolled-DEQ-iPPG (UDEQ-iPPG)}. In each unrolling iteration, Unrolled-DEQ-iPPG effectively applies the denoising operator on the pulse signal until a fixed-point is achieved, but applies the operator on the noise only once through a single forward pass. We show that UDEQ-iPPG exceeds performance of both unrolling and fixed-point algorithms by integrating both techniques.

Extensive experiments demonstrate that our models accurately estimate the underlying pulse signal from facial video, achieving new state-of-the-art heart rate estimation results on three publicly available datasets, and even generalizing to unseen datasets at test time. 

To summarize, our main contributions are as follows:

\begin{itemize}
    \item We model iPPG pulse signal recovery and heart rate estimation as an optimization problem that minimizes the sum of a data fidelity term and an \textit{implicit} structure-promoting prior. We then learn the denoising neural network operator corresponding to the implicit prior.
    \item We present the first unrolling techniques for iPPG signal recovery. Using the ideas of proximal gradient descent, we alternate between gradient descent and neural network denoising on both the signal and noise to recover the pulsatile signal. This algorithm is called Unrolled iPPG.
    \item We are the first to propose iPPG signal recovery through fixed-point methods and to incorporate deep equilibrium models into algorithms for pulse signal recovery and heart rate estimation. We do so by encapsulating both the gradient and denoising steps in an iteration map that runs until a fixed-point is obtained for both the pulse and noise signals and call this DE-Prox-iPPG.
    \item We further propose UDEQ-iPPG, which integrates DE-Prox-iPPG and Unrolled iPPG. Assuming that the pulse signal can be represented as the fixed-point of a learned denoising operator, while the noise admits no such fixed-point representation,  we apply the denoiser on the pulse signal until fixed-point convergence, but we apply the denoiser on the noise a single time at each iteration in an unrolling framework.
    \item We achieve state-of-the-art heart-rate estimation results on three publicly available datasets, while using less than 35\% of the number of parameters of competing deep neural network methods.
\end{itemize}

A preliminary version of Unrolled iPPG appeared in~\cite{shenoyUnrolled2023}. We expand upon this work by: 1) Incorporating fixed-point methods for signal recovery (DE-Prox-iPPG); 2) showing that a combination of fixed-point recovery algorithms and traditional single-pass feed-forward neural networks effectively models the signal and noise (UDEQ-iPPG); 3) testing all three methods on two additional datasets; 4) performing cross-dataset model evaluations; and 5) providing a more thorough empirical and statistical analysis of all of our methods.

\section{Related Work}
\label{sec:related_works}
\subsection{Imaging PPG and Physiological Analysis}\label{subsec:remote_physiological_analysis}

Pulse rate estimation algorithms can be broadly categorized into signal-processing-based (or model-based) algorithms and deep learning algorithms. Blind source separation (BSS) methods~\cite{lewandowska,poh2010non, favilla2018heart}, a subset of the signal processing methods, assume that the intensity signal extracted from video is the weighted sum of a pulse signal and other source or error signals. By assuming the source signals are statistically independent, \cite{lewandowska} uses independent component analysis (ICA) to extract the pulse signal, while~\cite{poh2010non} demixes signals along directions of maximum variance using principal components analysis (PCA). These methods were further developed to include skin reflection properties. The method of~\cite{pbv} does so by assuming all color variation in skin pixels is due to the pulsatile signal, and projects the extracted time series onto this color vector, the resultant signal of which is proportional to the underlying pulse. CHROM~\cite{chrom} first implements a skin tone correction, followed by a projection that eliminates specular reflection to extract only the pulse signal. POS~\cite{wang2016algorithmic}, similar to CHROM, eliminates the specular reflection and performs skin tone correction, but defines a projection plane using a data-driven approach. Other methods have used low-rank~\cite{tulyakov2016self} and sparsity-based methods. The method of~\cite{autosparseppg} is most similar to our work; they define the extracted video signal as the sum of the pulse signal and noise, and assume the Fourier coefficients of the pulse signal and the noise signal to be sparse. They then solve a sparse optimization problem for the frequency coefficients and noise using proximal gradient descent. We propose to relax these sparsity assumptions by incorporating a learnable prior, which is realized through deep neural networks.

Deep learning methods \cite{Spetlik2018VisualHR, chen2018deepphys, nowara2021benefit, sun2022contrast, liu2021metaphys, yu2019physnet, liu2020tscan, liu2023efficientphys, yu2022physformer, yu2019remote, rppgDisentangle, bousefsaf20193d, hu2021eta, yu2020autohr, das2021bvpnet, lee2020meta, du2023dual} show significant performance improvements over model-based approaches. Many of these methods accept RGB frames directly; PhysNet~\cite{yu2019physnet} inputs RGB frames to spatiotemporal neural networks to reconstruct the pulse signal. DeepPhys~\cite{chen2018deepphys} uses the temporal difference of frames to eliminate the stationary skin reflection color and reconstruct the pulse wave. Both TS-CAN~\cite{liu2020tscan} and InverseCAN~\cite{nowara2021benefit} extend DeepPhys, with the former including temporal shift modules~\cite{lin2019tsm} and the latter estimating corruption in the background to remove noise in the foreground and improve signal-to-noise ratio (SNR). Some recent work evaluates algorithms under stricter data constraints such as the few-shot scenario~\cite{liu2021metaphys}, while EfficientPhys~\cite{liu2023efficientphys} develops pulse signal estimation algorithms assuming a resource-constrained environment. Recent work~\cite{Gideon_2021_ICCV, sun2022contrast, rppgDisentangle, liu2024rppg} has developed neural methods with no supervisory training signal, showing good performance.

\subsection{Deep Equilibrium Models} \label{subsec:implicit_models_in_deep_learning} 
Conventional deep learning models learn by applying explicit functions (such as linear and convolutional operators, followed by nonlinearities) to inputs, building a computation graph along the way, and back-propagating through this graph to update parameters after a loss computation. Implicit models such as Neural ODEs~\cite{neuralODE} and Deep Equilibrium Models~\cite{bai2019deep} do not build computation graphs; instead, they solve a nonlinear system prescribed by a criterion such as the endpoint of an ODE flow or the root of an equation, and backpropagate through this point. Thus, they decouple the choice of solver from the computation of the endpoint or root, and backpropagate through this point analytically and independently of the forward pass.

Deep Equilibrium Models (DEQs)~\cite{bai2019deep} approximate ``infinitely deep" models by solving for the fixed point (or equilibrium point) $\mathbf{z^*}$ of a function: $\mathbf{z^*} = f_\theta(\mathbf{z^*}, \mathbf{x})$, where $\mathbf{x}$ is the input. The choice of solver for $\mathbf{z^*}$---whether it be Broyden's Method~\cite{broyden}, Peaceman-Rachford splitting~\cite{peaceman_rachford}, Anderson Acceleration~\cite{anderson}, or any other method---is independent of the backpropagation procedure. Once this point is found, the Implicit Function Theorem~\cite{implicit_function} can be used to compute the gradients of the parameters. These implicit models have shown competitive performance on word-level language modeling~\cite{bai2019deep} as well as ImageNet classification and Cityscapes semantic segmentation~\cite{bai2020multiscale}. Further developments showed that these models can simultaneously perform forward inference as well as input optimization on problems such as generative modeling, adversarial training, and gradient-based meta-learning~\cite{gurumurthy2021joint}. Recent work has applied DEQs to optical flow estimation~\cite{deq-flow}, diffusion models~\cite{deq-diffusion}, facial landmark estimation~\cite{deq-landmark}, and imaging inverse problems~\cite{gilton-inverse, gkillas2023highly, equilibrium-image-denoising, online-deq,deq-measurement-model, deq-explicit-reg}. In addition, significant research has explored the theoretical foundations of DEQs. After~\cite{bai2019deep} demonstrated training instabilities in DEQs, inexact gradient methods~\cite{geng2021training, fung2021jfb} and regularization techniques~\cite{bai2021stabilizing} were developed that improved fixed-point convergence, while the work of~\cite{wei2021certified, cer-deq} explored interval bound propagation methods to compute lower and upper bounds on the output of a layer. Convergence of DEQs were studied \cite{pabbaraju2021estimating,winston-monotone} using monotone operator theory, while~\cite{gan2022selfsupervised} looked at self-supervised learning of DEQs from a learning theory perspective. 

We are the first to apply deep equilibrium models to heart rate estimation. We will show that these formulations achieve state-of-the-art results on standard public datasets.

\section{Signal Model}\label{sec:approach}
\begin{figure*}
    \centering
    \includegraphics[width=0.85\textwidth]{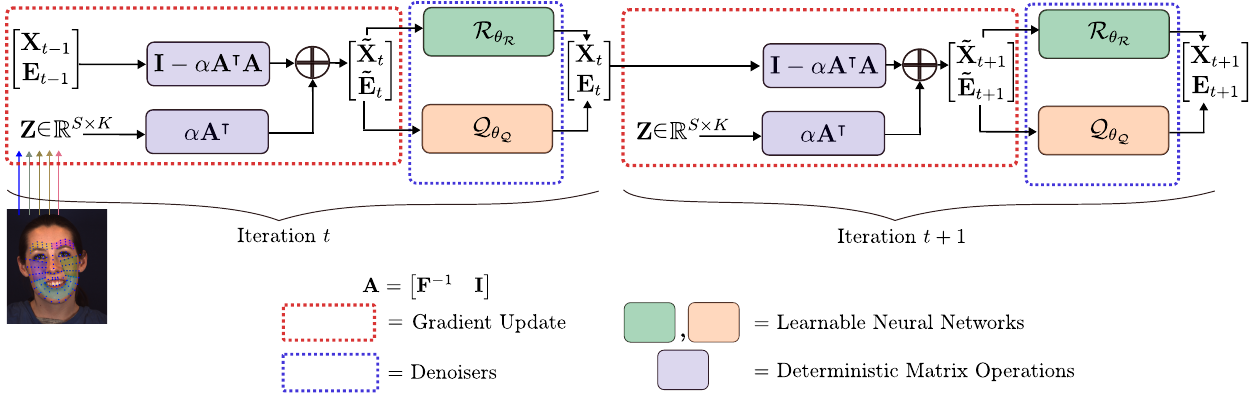}
    \caption{Our unrolling algorithm for both Unrolled iPPG and UDEQ-iPPG. We replace the proximal operators of proximal gradient descent by deep denoisers $\mathcal{R}_{\theta_\mathcal{R}}$ and $\mathcal{Q}_{\theta_\mathcal{Q}}$ on the frequency coefficients $\mathbf{X}$ and noise $\mathbf{E} $, respectively. Unrolled iPPG implements a single forward pass through $\mathcal{R}_{\theta_\mathcal{R}}$ and $\mathcal{Q}_{\theta_\mathcal{Q}}$, while UDEQ-iPPG effectively applies $\mathcal{R}$ to $\mathbf{\Tilde{X}}_t$ until the output converges to a fixed-point.}
    \label{fig:unrolling_algorithm}
\end{figure*}

The circulation of blood in the human body can be described as a pump---the heart---that rhythmically applies a force on a branched system of elastic tubes to move blood~\cite{nichols2022mcdonald}. Optical sensors such as those that perform photoplethysmography (PPG) capture blood volume changes by measuring transmission or reflection of illumination that is incident on the skin. While PPG measures this based on contact with the skin, we model the pulse signal based on non-contact imaging: reflected light from the skin that is captured via frames of an RGB camera.

Given a video segment with $S$ video frames, $C$ color channels, and height and width dimensions $H \times W$ pixels, assume that a time series $\mathbf{Z} \in \mathbb{R}^{S\times K}$ has been extracted from $K$ facial regions (described in Section~\ref{subsec:time_series_extraction}), where the set of $K$ time series is assumed to capture a combination of the pulsatile signal and noise:

\begin{equation}\label{eq:signal_model}
    \mathbf{Z} = \mathbf{F}^{-1}\mathbf{X} + \mathbf{E}.
\end{equation}
Here, $\mathbf{Z}\in \mathbb{R}^{S\times K}$ is a time series of length $S$ frames extracted from the image intensities of $K$ facial regions of the video, $\mathbf{F}^{-1} \in \mathbb{C}^{S\times N}$ is the oversampled inverse Fourier transform matrix, $\mathbf{X} \in \mathbb{C}^{N \times K}$ represents the frequency coefficients of the underlying pulse signal in the $K$ face regions, and \mbox{$\mathbf{E} \in \mathbb{R}^{S\times K}$} is a non-pulsatile noise matrix that captures noise due to multiple sources such as specular reflections, motion, and camera quantization error.

We aim to recover the pulse signal $\mathbf{X}$ and extract the noise $\mathbf{E}$ from the signal $\mathbf{Z}$. We model this as the minimization of the sum of a data fidelity term and regularizing penalty function:

\begin{equation}\label{eq:signal_recovery}
    \begin{aligned}
        \min_{\mathbf{X}, \mathbf{E}} \underbrace{\frac{1}{2}\norm{\mathbf{Z - A} \begin{bmatrix}
             \mathbf{X}\\
            \mathbf{E} 
         \end{bmatrix}}_F^2}_{\text{Data fidelity } = D} \; + \; \lambda\cdot \underbrace{\rho(\mathbf{X}, \mathbf{E})}_{\text{Regularization}},
    \end{aligned}
\end{equation}

where $\mathbf{A} = [\mathbf{F}^{-1} \quad \mathbf{I}]$. The data fidelity term, $D$, ensures consistency of the recovered signal with the measurements, while the regularization term imposes a prior on the signal and noise components. This optimization problem can be solved via proximal gradient descent, which alternates between a gradient update with respect to $D$ and a proximal mapping operation corresponding to the regularizer $\rho(\mathbf{X}, \mathbf{E})$, such that,

\begin{equation}\label{eq:signal_recovery_algorithm}
    \begin{split}
        \begin{bmatrix}
            \mathbf{\Tilde{X}}_{t+1} \\
            \mathbf{\Tilde{E}}_{t+1}
         \end{bmatrix} &\stackrel{\text{(a)}}{=} \begin{bmatrix}
             \mathbf{X}_{t} \\
            \mathbf{E}_t 
         \end{bmatrix} - \alpha 
             \nabla_{ \begin{psmallmatrix}
                 \mathbf{X} \\ \mathbf{E}
             \end{psmallmatrix}} D \left(\begin{bmatrix}
             \mathbf{X}_{t} \\
            \mathbf{E}_t 
         \end{bmatrix}\right) \\
         &\stackrel{\text{(a)}}{=} \alpha\mathbf{A}^\mathsf{T} \mathbf{Z} + (\mathbf{I} - \alpha\mathbf{A^\mathsf{T}A})\begin{bmatrix}
             \mathbf{X}_t \\
            \mathbf{E}_t
         \end{bmatrix},\\         
             \begin{bmatrix}
             \mathbf{X}_{t+1} \\
            \mathbf{E}_{t+1} 
         \end{bmatrix} & \stackrel{\text{(b)}}{=} \text{prox}_{\lambda\rho}\left(\begin{bmatrix}
            \mathbf{\Tilde{X}}_{t+1} \\
            \mathbf{\Tilde{E}}_{t+1}
         \end{bmatrix}\right),
    \end{split}
    \end{equation}
where ``(a)" defines the gradient step and ``(b)" defines the proximal step for an iteration $t$. The proximal operator is defined as

\begin{equation}\label{eq:proximal_def}
    \text{prox}_f(v) = \argmin_x\left( f(x) + \frac{1}{2}\norm{x -v}_2^2\right).
\end{equation}

When $\rho(\cdot, \cdot)$ is explicitly defined, we may be able to calculate a closed form solution for the proximal operator. In this paper, we instead define the regularization term \textit{implicitly}, and we present three paradigms that instantiate the proximal operators as denoising neural networks. 

\section{Algorithms}\label{sec:Algorithms}

Unlike previous work~\cite{autosparseppg}, which explicitly defines the regularizer $\rho(\mathbf{X}, \mathbf{E})$ as in~\eqref{eq:signal_recovery}, we assume that the penalty term is a complicated function that can not be written in closed form. In the framework of proximal gradient descent, $\rho(\mathbf{X}, \mathbf{E})$ is not minimized directly, but rather the minimization is realized through its proximal operator as shown in Equation~\eqref{eq:proximal_def}. In this paper, we approximate the proximal operator as a deep denoiser, proposing three methods: an unrolling algorithm (Section~\ref{subsec:unrolled-ippg}), a fixed-point algorithm (Section~\ref{subsec:prox-deq_ippg}), and a combination of unrolling and fixed-point algorithms (Section~\ref{subsubsec:udeq_desc}).

\subsection{Unrolling Algorithm: Unrolled iPPG}\label{subsec:unrolled-ippg}
Previous work~\cite{autosparseppg} defined explicit sparsity priors on the frequency coefficients of the pulse signal, making an assumption that the frequency coefficients are independent of each other. However, we hypothesize that sparsity (or joint sparsity) is not sufficiently expressive to characterize the inherent structure of the heartbeat signal. Rather, we explore how the complex, nonlinear combination of the frequency coefficients can be learned from data. Maintaining the algorithmic structure of proximal gradient descent, we  ``unroll"~\cite{monga-unrolling} the iteration map from Equation~\eqref{eq:signal_recovery}

\begin{equation}\label{eq:dp_ippg_algorithm}
    \begin{split}
        \begin{bmatrix}
            \mathbf{\Tilde{X}}_{t+1} \\
            \mathbf{\Tilde{E}}_{t+1}
         \end{bmatrix} & \stackrel{\text{(a)}}{=} \begin{bmatrix}
             \mathbf{X}_t \\
            \mathbf{E}_t 
         \end{bmatrix} - \alpha 
             \nabla_{ \begin{psmallmatrix}
                 \mathbf{X}_t \\ \mathbf{E_t}
             \end{psmallmatrix}} D \left(\begin{bmatrix}
             \mathbf{X}_t \\
            \mathbf{E}_t 
         \end{bmatrix}\right), \\         
             \mathbf{X}_{t+1} & \stackrel{\text{(b)}}{=} \mathcal{R}_{\theta_{\mathcal{R}}}(\mathbf{\Tilde{X}}_{t+1}) \\
            \mathbf{E}_{t+1} & \stackrel{\text{(b)}}{=} \mathcal{Q}_{\theta_{\mathcal{Q}}}(\mathbf{\Tilde{E}}_{t+1})
    \end{split}
\end{equation}
 where $\mathcal{R}_{\theta_{\mathcal{R}}}(\mathbf{\Tilde{X}}_{t+1})$ and $\mathcal{Q}_{\theta_{\mathcal{Q}}}(\mathbf{\Tilde{E}}_{t+1})$ are denoising functions implemented as neural networks in step (b). We graphically show the algorithm in Figure~\ref{fig:unrolling_algorithm}. Similar to proximal gradient descent, the gradient descent step ensures the recovered signal's consistency with the measurements, while the denoising operators impose the learned signal prior. We unroll for $T$ iterations, after which we compute a loss between the predicted signal and ground-truth (during training), or output the estimated pulse signal (during inference).

\subsection{Fixed-Point Algorithm: DE-Prox-iPPG}\label{subsec:prox-deq_ippg}

\begin{figure*}
    \centering
    \includegraphics[width=0.65\textwidth]{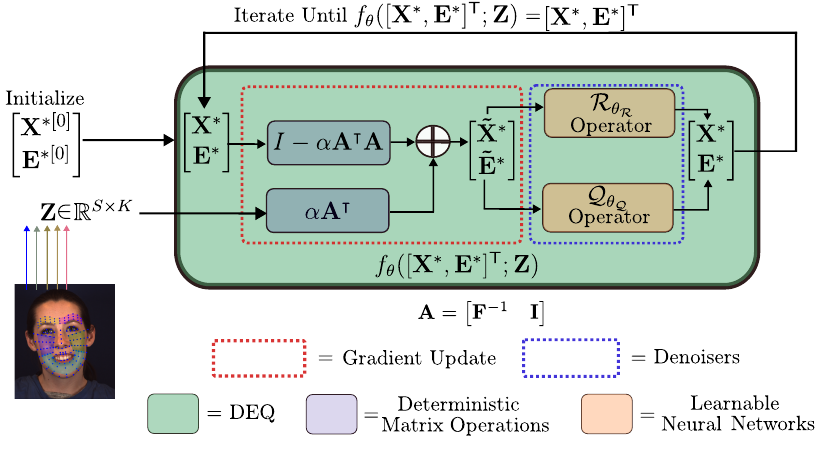}
    \caption{Our DE-Prox-iPPG algorithmic paradigm. The video-extracted time series $\mathbf{Z}$ and the two concatenated optimization variables $[\mathbf{X}^*,\mathbf{E}^*]^\mathsf{T}$ are passed through the block outlined in red which performs gradient updates $\frac{\partial \mathbf{D}}{\partial \mathbf{X}^*}, \frac{\partial \mathbf{D}}{\partial \mathbf{E}^*}$. The results $\mathbf{\Tilde{X}}^*$ and $\mathbf{\Tilde{E}}^*$ are then passed through denoising operators $\mathcal{R}$ and $\mathcal{Q}$. This process is repeated until fixed-point convergence.}
    \label{fig:prox_deq_pipeline}
\end{figure*}

A potential drawback of Unrolled iPPG is that the gradient and denoising steps are applied for a predetermined number of iterations $T$, where $T$ is a hyperparameter. It may be possible to apply these steps until convergence instead, obviating the need to choose $T$. Noting that proximal gradient descent can be viewed as a fixed-point iteration~\cite{parikhProximl, gilton-inverse}, we propose Deep Equilibrium Proximal Gradient Descent iPPG (DE-Prox-iPPG). This method, illustrated in Figure~\ref{fig:prox_deq_pipeline}, encapsulates the entire iteration map (i.e., gradient descent followed by denoising) in a Deep Equilibrium model (DEQ) and iterates these steps until a fixed point is found for both optimization variables. Compared to unrolling methods, we assume that we can find a joint fixed point for the frequency coefficients $\mathbf{X}^*$ and noise $\mathbf{E}^*$. Therefore, we optimize for the fixed point of the concatenation of the two variables $\mathbf{X}^*$ and $\mathbf{E}^*,$ and define a deep equilibrium model $f_\theta(\cdot, \cdot)$ that iterates through (a) gradient descent steps and (b) denoising operators until convergence:

 \begin{equation}\label{eq:deprox_algorithm}
    f_\theta\left(\begin{bmatrix}
            \mathbf{X}^* \\
            \mathbf{E}^*
         \end{bmatrix}, \mathbf{Z}\right) = 
     \begin{cases} 
      \begin{bmatrix}
            \mathbf{\Tilde{X}}^* \\
            \mathbf{\Tilde{E}}^*
         \end{bmatrix}
         & \stackrel{\text{(a)}}{=} \begin{bmatrix}
             \mathbf{X}^* \\
            \mathbf{E}^* 
         \end{bmatrix} - \alpha 
             \nabla_{ \begin{psmallmatrix}
                 \mathbf{X}^* \\ \mathbf{E^*}
             \end{psmallmatrix}} D \left(\begin{bmatrix}
             \mathbf{X}^* \\
            \mathbf{E}^* 
         \end{bmatrix}\right)\\         
             \mathbf{X}^* & \stackrel{\text{(b)}}{=} \mathcal{R}_{\theta_{\mathcal{R}}}\big(\mathbf{\Tilde{X}}^*\big), \\
            \mathbf{E}^* & \stackrel{\text{(b)}}{=} \mathcal{Q}_{\theta_{\mathcal{Q}}}\big(\mathbf{\Tilde{E}}^*\big).
   \end{cases}
\end{equation}

\noindent DE-Prox-iPPG computes the gradient and denoising steps until a fixed point of the \textit{global system} is found. While Unrolled iPPG unrolls for $T$ iterations, the number of DE-Prox-iPPG iterations varies depending on the input window, as well as on the joint fixed-point of the signal and noise variables.

 \subsection{Unrolling + Fixed-Point Algorithm: UDEQ-iPPG}\label{subsubsec:udeq_desc}

While DE-Prox-iPPG computes the fixed-point iteration for both the pulse signal and noise, we recognize that fixed-point optimization for both signals may be suboptimal. To  address this, we propose Unrolled-DEQ-iPPG (UDEQ-iPPG), which is based on the following two assumptions: 1) Given that the pulsatile signal is a realization of the quasi-periodic nature of the beating heart, we assume it is sufficiently structured to be represented as a fixed point of some neural network-based denoising operator; and 2) we assume that the variation in the source of the noise, such as motion or illumination, \textit{does not} admit a structure that would be appropriately represented as the fixed point of an operator. To that end,  we instantiate $\mathcal{R}$ in Figure~\ref{fig:unrolling_algorithm} as a DEQ but restrict $\mathcal{Q}$ to a single forward pass of a neural network, and we embed both operators in the Unrolled iPPG structure. The corresponding iteration map is summarized as follows:

 \begin{equation}\label{eq:udeq_algorithm}
    \begin{split}
        \begin{bmatrix}
            \mathbf{\Tilde{X}}_{t+1} \\
            \mathbf{\Tilde{E}}_{t+1}
         \end{bmatrix} & \stackrel{\text{(a)}}{=} \begin{bmatrix}
             \mathbf{X}_{t} \\
            \mathbf{E}_t 
         \end{bmatrix} - \alpha 
             \nabla_{ \begin{psmallmatrix}
                 \mathbf{X} \\ \mathbf{E}
             \end{psmallmatrix}} D \left(\begin{bmatrix}
             \mathbf{X}_{t} \\
            \mathbf{E}_t 
         \end{bmatrix}\right), \\
         & \stackrel{\text{(a)}}{=} \alpha\mathbf{A}^\mathsf{T} \mathbf{Z} + (\mathbf{I} - \alpha\mathbf{A^\mathsf{T}A})\begin{bmatrix}
             \mathbf{X}_t \\
            \mathbf{E}_t
         \end{bmatrix},\\         
             \mathbf{X^*} & \stackrel{\text{(b)}}{=} \mathcal{R}_{\theta_{\mathcal{R}}}(\mathbf{X^*};\mathbf{\Tilde{X}}_{t+1}) \\
        \mathbf{X}_{t+1} &= \mathbf{X^*}, \\
            \mathbf{E}_{t+1} & \stackrel{\text{(b)}}{=} \mathcal{Q}_{\theta_\mathcal{Q}}(\mathbf{\Tilde{E}}_{t+1}).
    \end{split}
    \end{equation}

\begin{figure}
    \centering
    \includegraphics[width=0.40\textwidth]{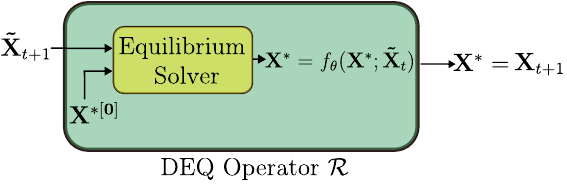}
    \caption{The DEQ operator used in UDEQ-iPPG. At each unrolling iteration, we apply the denoising operator until a fixed-point is found.
    }
    \label{fig:deq_only}
\end{figure}

The DEQ operator $\mathcal{R}(\cdot, \cdot)$ for UDEQ-iPPG is illustrated in Figure~\ref{fig:deq_only}. In each unrolling iteration, we input the gradient-updated frequency coefficients $\mathbf{\Tilde{X}}_{t+1}$ to the DEQ to produce the fixed-point variable $\mathbf{X}^*$ using a fixed-point solver. On the other hand, only a single forward pass through $\mathcal{Q}$ occurs for the noise variable $\mathbf{\Tilde{E}}$. The overall procedure is unrolled over $T$ iterations, similar to Unrolled iPPG.

\subsection{Parameter Updates for Fixed-Point Networks}

The fixed point in DE-Prox-iPPG  can be found na{\"i}vely by repeated application of $f_{\theta}(\cdot, \cdot) $  until $\big\|\big[\begin{smallmatrix}
            \mathbf{X^*}\\
            \mathbf{E^*}
         \end{smallmatrix}\big] - f_{\theta}(\big[\begin{smallmatrix}
            \mathbf{X^*}\\
            \mathbf{E^*}
         \end{smallmatrix}\big];\mathbf{Z}) \big\|$ 
is below some tolerance. This method, unfortunately, builds large computation graphs in memory. Equivalently, we can find the root of 
         \begin{equation}
             g\left(\begin{bmatrix}
            \mathbf{X^*}\\
            \mathbf{E^*}
         \end{bmatrix};\mathbf{Z}\right) = \mathbf{X^*} - f_{\theta}\left(\begin{bmatrix}
            \mathbf{X^*}\\
            \mathbf{E^*}
         \end{bmatrix};\mathbf{Z}\right)  = \mathbf{0}.
         \end{equation}
         
Similarly, for UDEQ-iPPG, we would like to find the root of $h(\mathbf{X^*},\mathbf{\Tilde{X}}_{t+1})) = \mathbf{X^*} - f_{\theta} (\mathbf{X^*},\mathbf{\Tilde{X}}_{t+1})  = \mathbf{0}.$  This technique permits efficient root-solving algorithms such as Broyden's Method~\cite{broyden} or Anderson Acceleration~\cite{anderson} to solve for the fixed point. 

During training, we iterate for $T$ iterations for UDEQ-iPPG, or until fixed-point convergence for DE-Prox-iPPG, as shown in Figures \ref{fig:unrolling_algorithm} and \ref{fig:prox_deq_pipeline}, after which both models' parameters must be updated after computing some loss $\mathcal{L}$. While $\theta_{\mathcal{Q}}$ (denoting the parameters of $\mathcal{Q}$) in UDEQ-iPPG can be updated via the standard auto-differentiation modules, updating the parameters $\theta_{\mathcal{R}}$ in UDEQ-iPPG and both $\theta_{\mathcal{R}},\theta_{\mathcal{Q}}$ in DE-Prox-iPPG requires computing gradients with respect to the parameters. While a na{\"i}ve implementation would backpropagate through iterations of the solver, a memory-efficient backpropagation scheme uses the Implicit Function Theorem (see~\cite{bai2019deep} for a proof), shown below.

\begin{theorem}\label{thm:ift}
        {\rm Implicit Function Theorem (IFT) \cite{bai2019deep, implicit_function, deq-flow}}. Given the fixed-point representation $\mathbf{p}^*$ of the optimzation variables,  and the corresponding loss $\mathcal{L}(\mathbf{Z}, \mathbf{Z}_{\textup{gt}})$, where $\mathbf{Z} = \mathbf{F^{-1}X^*}$, the gradient of the DEQ flow is given by

    \begin{equation}
        \frac{\partial \mathcal{L}}{\partial \theta} = \frac{\partial \mathcal{L}}{\partial \mathbf{p^*}}
        \left(\mathbf{I} - \frac{\partial f_{\theta}(\mathbf{p^*})}{\partial \mathbf{p^*}}\right)^{-1} \frac{\partial f_{\theta}(\mathbf{p^*, Z})}{\partial \theta}.
    \end{equation}
\end{theorem}

Note that for UDEQ-iPPG, $\mathbf{p}^* = \mathbf{X}^*$, while for DE-Prox-iPPG it is $\mathbf{p}^* = \big[\begin{smallmatrix}             \mathbf{X^*}\\ \mathbf{E^*}\end{smallmatrix}\big] $. We compute the mean squared error (MSE) Loss between the final output $\mathbf{Z} = \mathbf{F^{-1}X^*}$ and the ground-truth PPG signal $\mathbf{Z}_{\text{gt}}$. Here, $\mathbf{X}^*$ is $\mathbf{X}^*_T$ in UDEQ-iPPG (i.e. the output of the fixed-point operator at iteration $T$)  and is $\mathbf{X}^*$ (the fixed point) in DE-Prox-iPPG.  To ensure stability of the solution, we add a regularizing penalty function given by the Frobenius norm of the Jacobian with respect to the fixed-point $\mathbf{p^*}$, which is given by $\mathbf{J}_{f_{\theta}} = \frac{\partial f_{\theta}(\mathbf{p^*})}{\partial \mathbf{p^*}},$ where $\mathbf{J}_{f_{\theta}} \in \mathbb{R}^{d\times d}$ and $d$ is the length of the vector $\mathbf{p}^*$~\cite{geng2021training,bai2021stabilizing}. To compute $\norm{\mathbf{J}_{f_{\theta}}}_F = ( \text{tr}(\mathbf{J}_{f_{\theta}}\mathbf{J}_{f_{\theta}}^\top))^{1/2}$, we use the Hutchinson estimator~\cite{hutchinson}: 

\begin{equation}\label{eq:hutchinson_estimator}
    \text{tr}(\mathbf{J}_{f_{\theta}}\mathbf{J}_{f_{\theta}}^\top) = \mathbb{E}_{\epsilon\in \mathcal{N}(\mathbf{0}, \mathbf{I}_d)} \left[ \norm{\epsilon^\mathbb{\top}\mathbf{J}_{f_{\theta}}}_2^2 \right]
\end{equation}

  Our final loss is composed of the mean squared error (MSE) between the recovered waveform and the ground-truth waveform plus the Jacobian regularization penalty:

\begin{equation}\label{eq:deq_loss_function}
    \mathcal{L} = \mathcal{L}_{\text{mse}} (\mathbf{Z}_{\text{final}},\mathbf{Z}_{\text{gt}}) + \lambda \frac{\norm{\epsilon^\mathbb{\top}\mathbf{J}_{f_{\theta}}(\mathbf{p^*})}_2^2}{d}, \epsilon \in \mathcal{N}(0, \mathbf{I}_d) ,
\end{equation}

where $\lambda$ determines the strength of regularization. 

\subsection{Pulse Rate Computation}

The output of each of the aforementioned algorithms are the Fourier coefficients $\mathbf{X} \in \mathbb{C}^{N \times K}$ of the $K$ regions from which we extracted a signal, where $N$ is number of frequency bins. To find the pulse rate, we sum the power of the frequency coefficients across all regions for each bin:
      
\begin{equation}
    \begin{split}
        r_n &= \sum_{k=1}^K |\mathbf{X}_{n, k}|^2 ,
    \end{split}
\end{equation}        
and the pulse rate is determined by the frequency bin that has the highest magnitude.

\section{Implementation and Experimental Results}\label{sec:expeirments}

\begin{figure}
\begin{tabular}{c@{\hspace{3pt}}c@{\hspace{3pt}}c}
\subfloat{\includegraphics[width=0.14\textwidth]{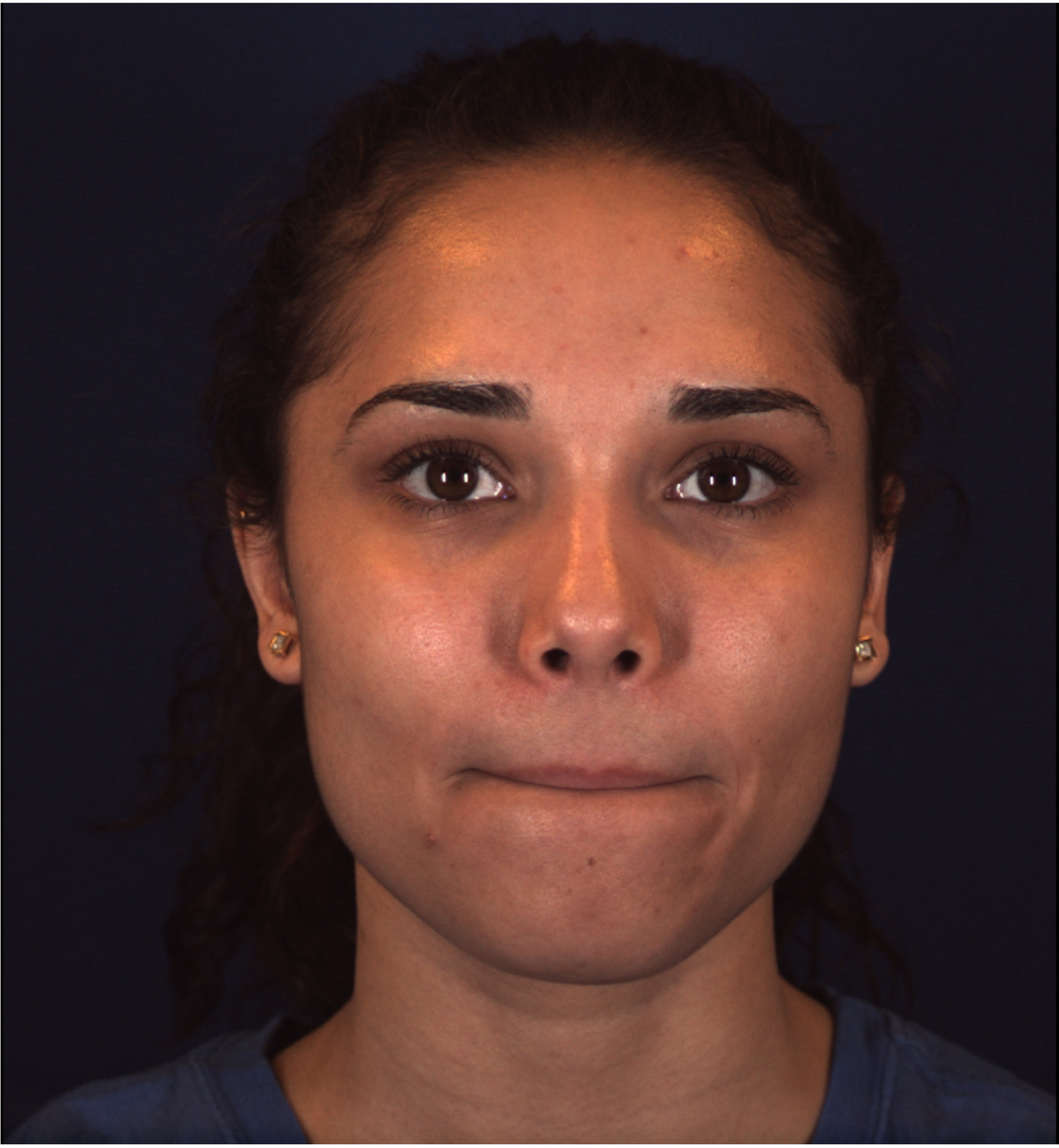}} &
\subfloat{\includegraphics[width=0.14\textwidth]{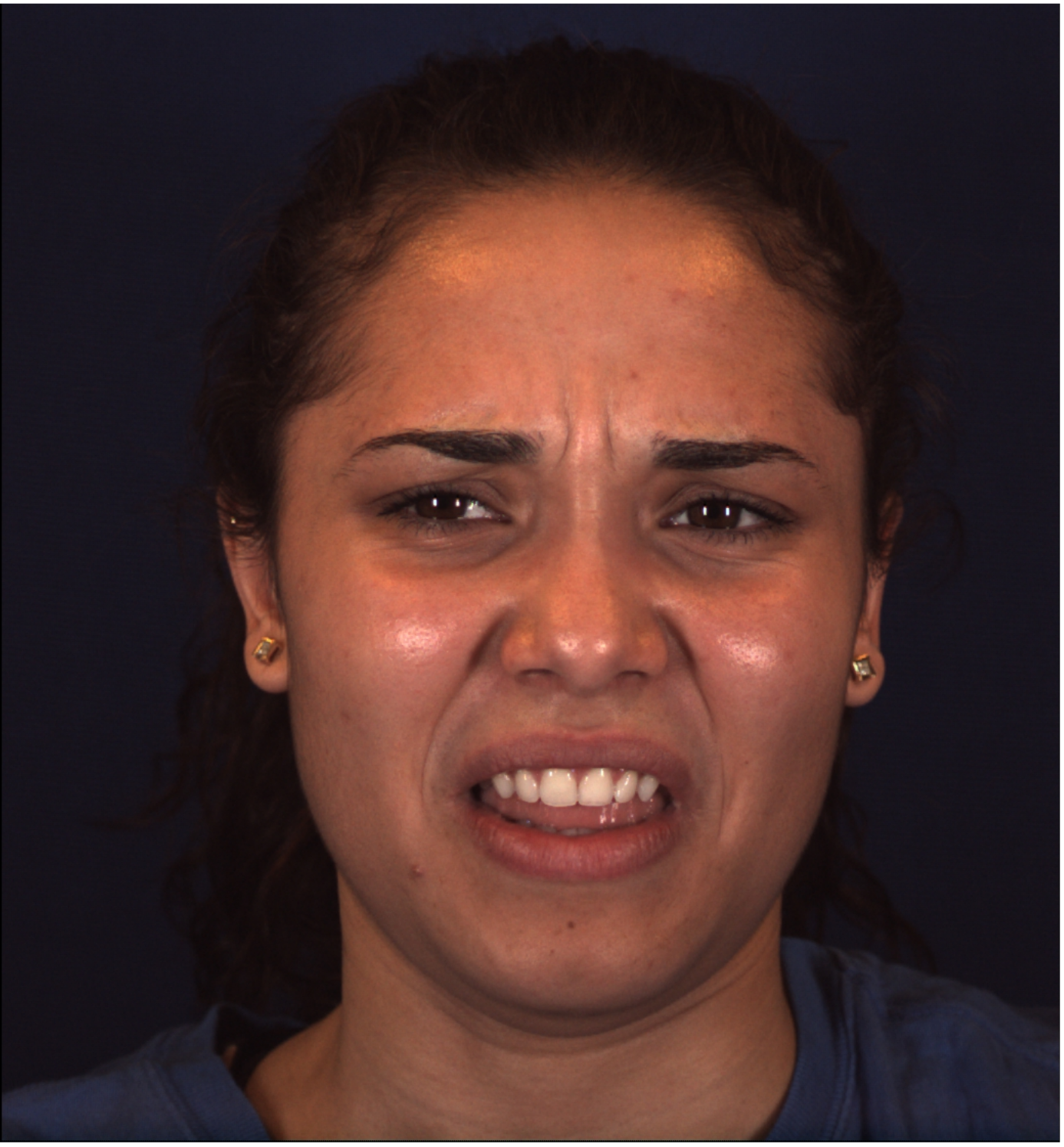}} &
\subfloat{\includegraphics[width=0.14\textwidth]{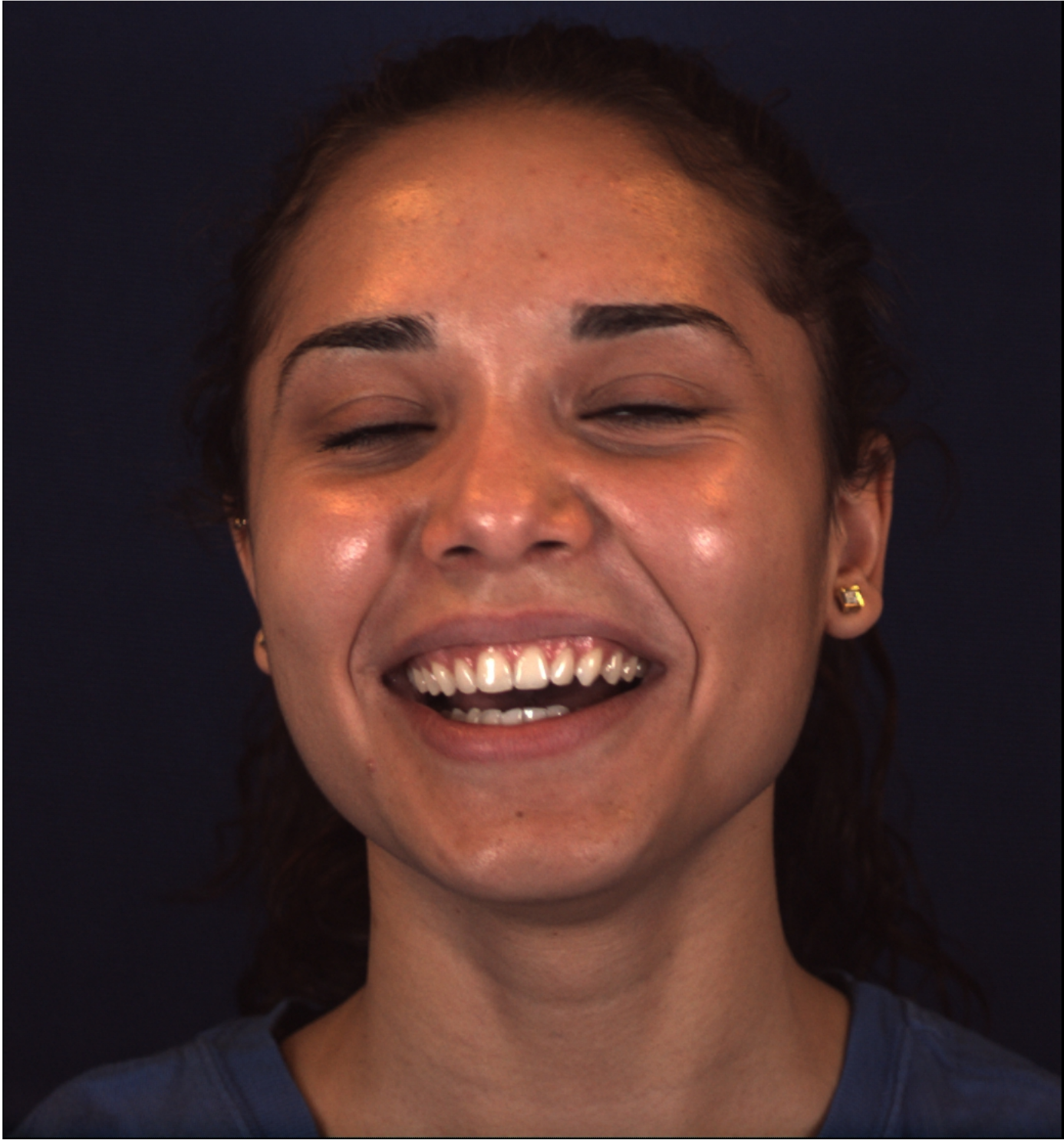}}\\
\subfloat{\includegraphics[width=0.14\textwidth]{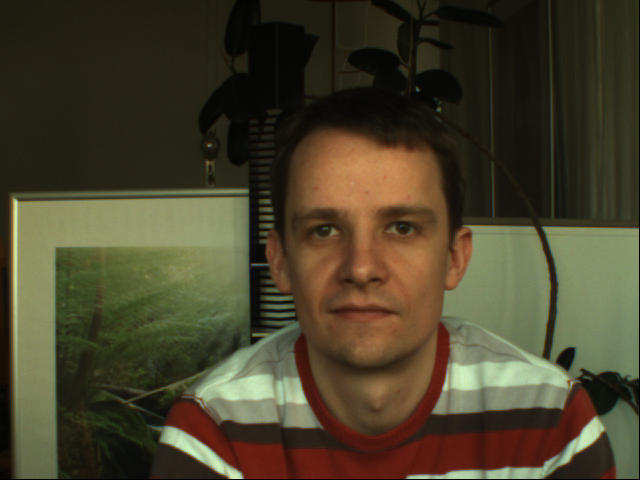}} &
\subfloat{\includegraphics[width=0.14\textwidth]{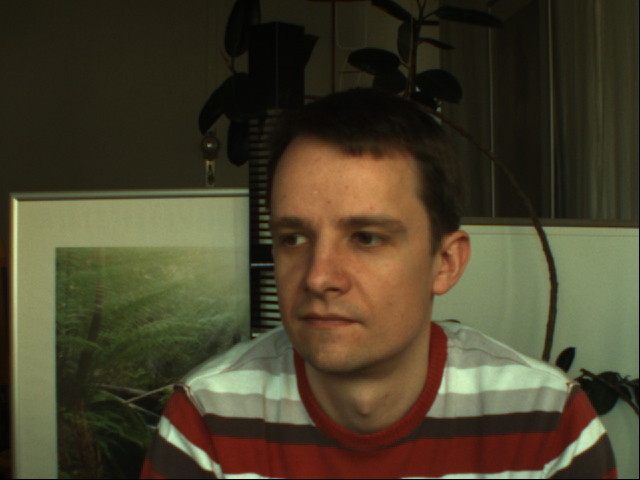}} &
\subfloat{\includegraphics[width=0.14\textwidth]{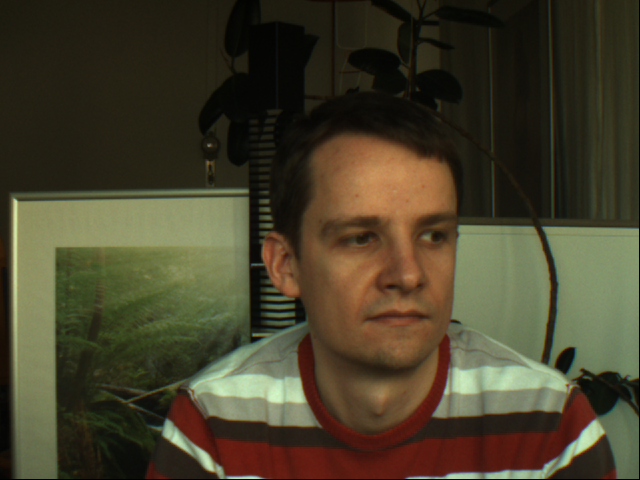}}\\[2pt]
\subfloat{\includegraphics[width=0.14\textwidth]{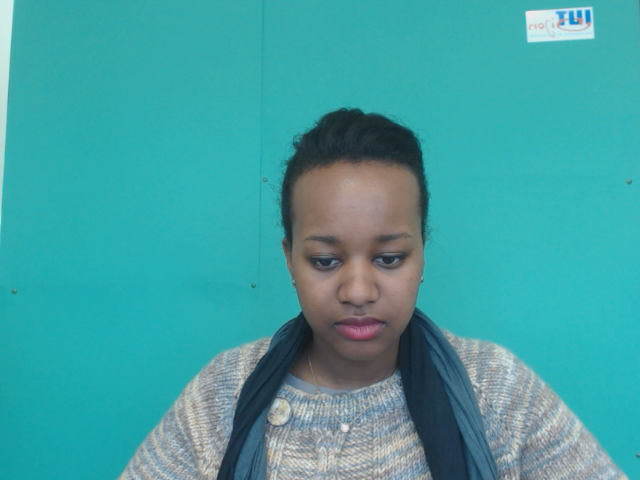}} &
\subfloat{\includegraphics[width=0.14\textwidth]{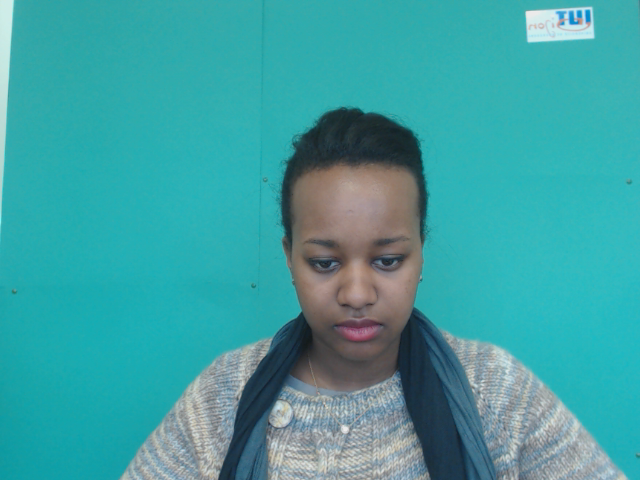}} &
\subfloat{\includegraphics[width=0.14\textwidth]{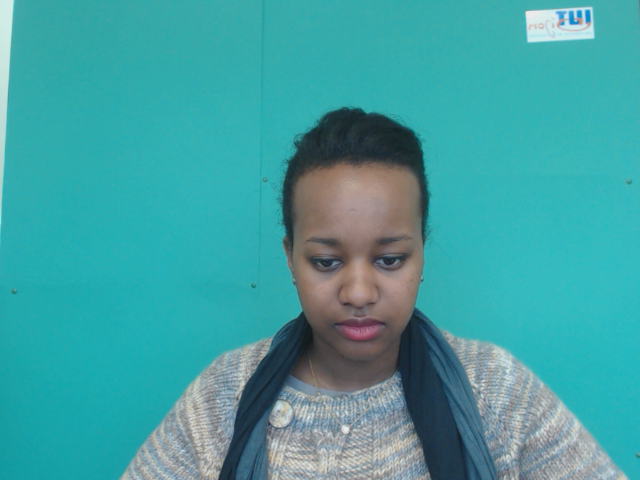}}

\end{tabular}
\caption{Image stills from the MMSE-HR dataset (top), PURE dataset (middle), and UBFC-rPPG dataset (bottom). }
    \label{fig:image_stills}
\end{figure}

\begin{figure}
    \centering
    \includegraphics[width=0.49\textwidth]{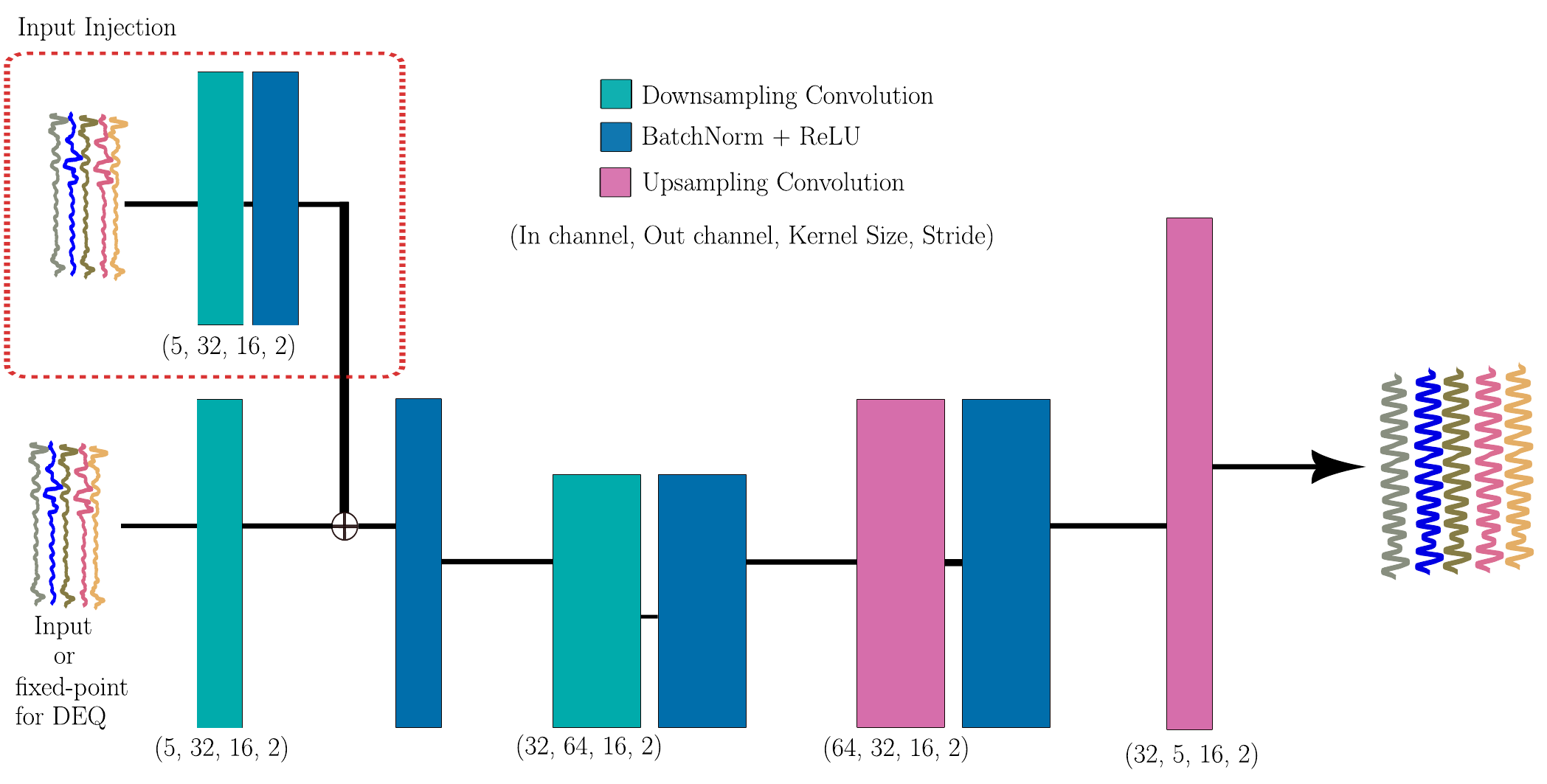}
    \caption{The architecture diagram of both $\mathcal{R}$ (complex-valued) and $\mathcal{Q}$ (real-valued). }
    \label{fig:architecture}
\end{figure}

\begin{table*}[]
    \caption{Heart rate estimation results on MMSE-HR, PURE and UBFC-rPPG datasets. Best results in each column are \textbf{bold}; second-best are \underline{underlined}.}

    \resizebox{0.98\textwidth}{!}{
    \begin{tabular}{|c|c|ccc|ccc|ccc|c|}
    \hline
    \multirow{3}{*}{\textbf{Type}} & \multirow{3}{*}{\textbf{Method}} & \multicolumn{3}{c|}{MMSE-HR} &  \multicolumn{3}{c|}{PURE} & \multicolumn{3}{c|}{UBFC-rPPG} & \textbf{No. Params }\\
    & & \textbf{MAE}$\downarrow$ & \textbf{RMSE}$\downarrow$& $\rho$ $\uparrow$ &  \textbf{MAE}$\downarrow$ & \textbf{RMSE}$\downarrow$& $\rho$ $\uparrow$ &  \textbf{MAE} $\downarrow$ & \textbf{RMSE}$\downarrow$& $\rho\uparrow$ & ($\times 10^5$) $\downarrow$\\
    \hline
    
    \multirow{4}{*}{Model-based} & ICA \cite{poh2010non}  & 5.44 & 12.00 & - &  - & - & - & 7.50 & 14.41 & 0.62 & - \\
    & CHROM \cite{chrom} & 3.74 & 8.11 & 0.55 & 2.07 & 9.92 & - & 2.37 & 4.91 & 0.89 & -\\
    & POS \cite{wang2016algorithmic} & 3.90 & 9.61 & - & 5.44 & 12.00 & - & 4.05 & 8.75 & 0.78  & -\\
    & AutoSparsePPG('20)\cite{autosparseppg}  & 4.55 & 14.42 & - & - & - & - & - & - & - & -\\
    \hline
    \multirow{12}{*}{Data-driven} & HR-CNN('18) \cite{Spetlik2018VisualHR} & - & - & - & 1.84 & 2.37 & - & - & - & - & 17.4 \\
    & SynRhythm('18) \cite{synrhythm} & - & - & - & - & - & - & 5.59 & 6.82 & 0.72 & -\\    
    & CAN('18) \cite{chen2018deepphys} & 4.06 & 9.51 & - & - & - & - & - & - & - & 5.3\\
    & CVD ('20)~\cite{niu2020video} & - & 6.04 & 0.84 & 1.29 & 2.01 & 0.98 &  2.19 & 3.12 & 0.99 &  123.3 \\
    & PulseGAN('21) \cite{SongPulseGAN} & - & - & -  & - & - & - & 1.19 & 2.10 & 0.98 & 7.6\\
    & InverseCAN('21) \cite{nowara2021benefit} & 2.27 & 4.90 & - & - & - & - & - & - & - & 10.7\\
    & Gideon('21)~\cite{Gideon_2021_ICCV} & - & - & - & 2.10 & 2.60 & 0.99 & 3.60 & 4.60 & 0.95 & 8.5 \\
    
    & DualGAN('22) \cite{dualganLu2021} & - & - & - & 0.82 & 1.31 & 0.99& \textbf{0.44} & \textbf{0.67} & \textbf{0.99} & 76.8\\
    & Physformer('22)~\cite{yu2022physformer} & 2.84 & 5.36 & - & - & - & - & - & - & - & 73.8 \\
    & Federated('22)~\cite{liu2022federated} & 2.99 & - & 0.79 & - & - & - & 2.00 & 4.38 & 0.93 & 5.3 \\
    & EfficientPhys-C('23)~\cite{liu2023efficientphys} & 2.91 & 5.43 & 0.86 & - & - & - & - & - & - & 4.6 \\
    & Contrastphys+('23)~\cite{sun2022contrast} & 1.11 & 3.83 & 0.96 & 0.48 & 0.98 & 0.99 & 0.50 & 0.84 &  \textbf{0.99} & 8.5\\
    
    \multirow{3}{*}{Hybrid} & \textbf{Unrolled iPPG (Ours)} & \underline{1.11} & 2.97 & 0.97 & 0.41 & 0.88 & 0.99 & 0.92 & 1.84 & 0.98 & \textbf{1.4}\\
    & \textbf{DE-Prox-iPPG (Ours)} & 1.22 & \underline{2.80} & \underline{0.95} & \textbf{0.13} & \textbf{0.28} & 0.99 & 2.86 & 3.84 & 0.97 & \textbf{1.4}\\
    & \textbf{UDEQ-iPPG (Ours)} & \textbf{0.95} & \textbf{2.53} & \textbf{0.98} & \underline{0.18} & \underline{0.35} & 0.99 & \underline{0.46} & \underline{0.82} & \textbf{0.99} & \textbf{1.4}\\
    \hline
    \end{tabular}}
    
    \label{tab:allresults}
\end{table*}

\begin{figure*}
    \centering
    \includegraphics[width=0.90\textwidth]{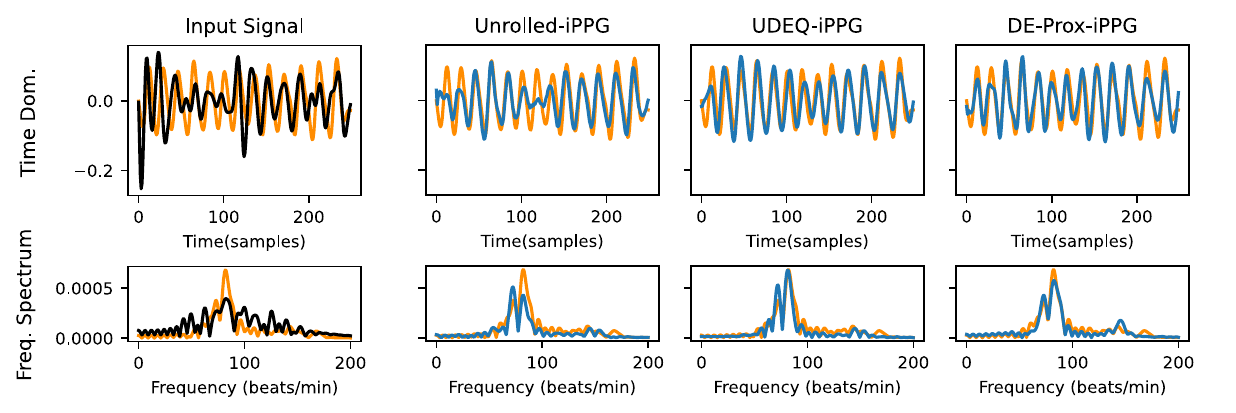}
    \caption{Comparing the input signal to results from Unrolled iPPG, UDEQ-iPPG, and DE-Prox-iPPG on the MMSE-HR dataset for one time window. The \textbf{\textcolor{orange}{orange}} waveform in all graphs is the ground truth pulse signal. The input signal (produced by our time-series extraction module) is shown in \textbf{black} in the first column, while the recovered signals output by our algorithms are shown in \textbf{\textcolor{blue}{blue}}. In each column, the top graph is plotted in the time domain, while the bottom graph shows the same plot in the frequency domain. It is clear from the frequency spectrum in the lower left that the \textbf{\textcolor{black}{input signal}} is noisy. All signal recovery algorithms output recover a time series and spectrum, but only UDEQ-iPPG and DE-Prox-iPPG recover the correct peak frequency.}
    \label{fig:comparison2}
\end{figure*}

\subsection{Datasets}\label{subsec:Datasets}
We evaluate our algorithm on three datasets, described below.

\textbf{MMSE-HR}~\cite{zhang2016multimodal-mmsehr, ertugrul2019cross}: The MMSE-HR dataset is a multimodal dataset with synchronized facial video and physiological monitoring. Facial video was captured at a resolution of 1040$\times$1392 at 25 fps (frames per second) using the Di3D dynamic imaging system, while the blood pressure waveform was captured by a finger cuff using the Biopac NIBP100D recording at 1000 Hz (downsampled to 25Hz) and calibrated using an arm cuff. A total of 102 videos from 17 male and 23 female subjects were captured as various emotions (happiness, surprise, embarrassment, fear, etc.) were elicited, resulting in substantial head motion.

We train on 10 s windows, shifting the time window by a stride of 2.4 s to obtain each new window.  For evaluation on the MMSE-HR dataset, we follow the evaluation protocol of \cite{comas2021turnip, nowara2021benefit} concatenating three 10 s windows to obtain a 30 s window for evaluation. We perform leave-one-subject-out cross-validation and evaluate on 30 s, non-overlapping time windows.

\textbf{PURE}~\cite{Stricker2014NoncontactVP}: The PURE dataset consists of 10 subjects who perform 6 different motion tasks, from ``no motion'' to ``fast translation'' and ``rotation''. The camera recorded at 30 fps at a resolution of 640$\times$480, while the corresponding pulse oximeter recorded at 60 Hz. The pulse oximeter was downsampled to match the rate of the video. We followed the train, validation, and test splits of~\cite{Spetlik2018VisualHR} to evaluate our algorithm, using 30 s, non-overlapping time windows as in~\cite{sun2022contrast} during test time. We train on 10 s windows, shifting the time window by a stride of 2.4 s to obtain each new window.

\textbf{UBFC-rPPG}~\cite{ubfc-rppg}: Captured using a webcam recording at 30 fps with a resolution of 640$\times$480, the UBFC-rPPG dataset contains 43 videos of subjects playing a time-sensitive mathematical game to induce heart rate changes. The pulse signal was captured using a pulse oximeter finger clip sensor recording at 30 Hz. To evaluate our algorithm on this dataset, we follow the protocols of~\cite{dualganLu2021}, which evaluated on non-overlapping 10 s time windows per video, and averaged these windows to get a single heart rate estimate for each video. They then computed the pulse rate metrics over the 12 pulse rate estimates (corresponding to the 12 videos in the test set). We train on 10 s windows, shifting the time window by a stride of 2.4 s to obtain each new window.

An example of the images from each dataset is in Figure~\ref{fig:image_stills}.

\subsection{Evaluation Metrics}\label{subsec:evaluation_metrics}
To compute the pulse rate, we first multiply our pulse signal window by a Hanning window. Next, letting $L = 100 \times \text{signal length}$, we compute the $L$-point FFT, and square the magnitudes of the frequency coefficients to get the power spectra. We sum the power for each frequency bin across all regions, and select the bin with the highest power as our pulse rate estimate. For pulse rate estimation, we report the mean absolute error (MAE) and root mean squared error (RMSE) between the predicted heart rate and ground-truth heart rate in a time window, measured in beats per minute as in~\cite{liu2022rppg}. We define the MAE and RMSE as:

\begin{equation}\label{eq:mae_calculation}
    \text{MAE} = \frac{1}{N} \sum_{i=1}^N |R_i - \hat{R}_i|, \text{RMSE} = \sqrt{\frac{1}{N} \sum_{i=1}^N (R_i - \hat{R}_i)^2} ,
\end{equation}

where $N$ is the number of time windows, $R_i$ is the ground-truth pulse-rate, and $\hat{R}_i$ is the predicted pulse rate in the time window. In addition, we measure the correlation coefficient $\rho$ between the predicted and ground-truth heart-rate estimates as in~\cite{liu2022rppg}.
As it is also important to know what percentage of our estimates are sufficiently close to the ground-truth pulse rates, we use the PTE6 metric~\cite{comas2021turnip}, defined as the percentage of the time that the estimated heart rate is within 6 bpm of the ground-truth heart rate: 

\begin{equation}\label{eq:pte6_calculation}
    \text{PTE}6 = \frac{100}{N} \sum_{i=1}^N P_i, \text{ where } P_i =  
\begin{cases}
    1,& \text{if } |R_i - \hat{R}_i| < 6\\
    0,              & \text{otherwise.}
\end{cases} 
\end{equation}

For a fair comparison, we use 30 s time windows for evaluation on the MMSE-HR and PURE datasets as in~\cite{sun2022contrast}, and 10 s time windows on the UBFC-rPPG dataset as in~\cite{dualganLu2021}. Note that the results for~\cite{sun2022contrast} on the UBFC-rPPG dataset were reevaluated to match the evaluation protocol of other methods such as~\cite{dualganLu2021} on UBFC-rPPG.
\subsection{Time Series Extraction}\label{subsec:time_series_extraction}

We generate the $S$-frame, $K$-channel time series 
\mbox{$\mathbf{Z}\in \mathbb{R}^{S\times K}$} from the video. To do so, we first detect the face using the FaceBoxes~\cite{Faceboxes} and crop the face; next, we pass these crops to the LUVLi landmark detection~\cite{kumar2020luvli} module, which generates 68 landmark points on the face as shown in Figure~\ref{fig:hr_pipeline}. To extract a pulsatile signal from the skin, we first interpolate new landmarks from the initial 68 landmarks to obtain an additional 77 landmarks that span the cheeks, chin, and forehead. We then incorporate two design choices: 1) We segment the face into $K = 5$ super-regions, and average each RGB color channel's pixel intensity values in these regions as described in~\cite{comas2021turnip}, and 2) we compute the ratio of the red color channel to green color channel as described in~\cite{chrom}. Both of these steps improve signal strength and reduce motion noise. Doing this for all frames, we obtain a time series $\mathbf{Z}\in \mathbb{R}^{S\times K}$.

After extracting the signals, each signal from the 5 spatial regions is convolved with a fifth-order Butterworth filter with cutoff frequencies of 0.7 Hz and 2.5 Hz as in~\cite{nowara2021benefit}. Finally, the signal is AC/DC normalized by dividing by the mean signal after subtracting the mean signal: $\hat{z}_k = \frac{z_k - \mu_k}{\mu_k}$, where $z_k$ is the signal in super-region $k$, and $\mu_k$ is the mean of that same signal. These signals are segmented into windows of $S$ samples and arranged as $\mathbf{Z}\in \mathbb{R}^{S\times K}$ for heart rate estimation.

\subsection{Implementation Details}\label{subsec:model_implementation_details}

The architecture diagram for $\mathcal{R}$ and $\mathcal{Q}$  is shown in Figure~\ref{fig:architecture}. Deep equilibrium models (DEQs) are instantiated as a collection of convolutions, linear operators, non-linearities, etc. organized as a single DEQ layer, and it is through this layer that the input is passed until convergence of a root-finding algorithm. To ensure a fair comparison, we adopt the same architecture throughout. The DEQ models contain an extra MLP as an input injection network to ensure stability during training~\cite{geng2021training,bai2021stabilizing}. Note that both $\mathcal{R}(\cdot)$ and $\mathcal{Q}(\cdot)$ have an identical architecture, with the only difference that the parameters of $\mathcal{R}(\cdot)$ are complex (since $\mathcal{R}(\cdot)$ operates in the frequency domain) while those of $\mathcal{Q}(\cdot)$ are real.

All three models are trained using a batch size of 100. On the MMSE-HR dataset, we train for 10 epochs using UDEQ-iPPG and 25 epochs with DE-Prox-iPPG,  with an initial learning rate for set to $3 \times 10^{-4}$ and reduced by half at the 10th epoch. For the PURE dataset, we train for 10 epochs on both models. On the UBFC dataset, we found using cross-validation that training for 45 epochs provided the optimal performance. The Jacobian regularization strength $\lambda$ in Equation~\eqref{eq:deq_loss_function} is set stochastically~\cite{bai2021stabilizing} during the training iterations using a random variable $x \sim \text{Bernoulli}(0.5)$, where $\lambda(x) = 5$ if $x=1$, otherwise $\lambda(x) = 0$. All parameters were determined by cross validation.

\subsection{Experimental Results}\label{subsec:experimental_results}
\subsubsection{Comparison to the State of the Art}

The results on the MMSE-HR, PURE, and UBFC-rPPG are shown in  Table~\ref{tab:allresults}. We see strong denoising performance for all three of our models.  Our UDEQ-iPPG model yields new state-of-the-art results on both the MMSE-HR and PURE datasets, exhibiting better performance than both purely deep learning algorithms such as \cite{sun2022contrast, liu2023efficientphys} and purely signal-processing, model-based methods such as \cite{autosparseppg, chrom, lewandowska}. On the UBFC-rPPG dataset, UDEQ-iPPG nearly matches the performance of DualGAN~\cite{dualganLu2021}, but does so with 50 times fewer trainable parameters. Our algorithms' strengths are shown qualitatively in Figure~\ref{fig:comparison2}, where the first column shows the input signals extracted from facial video in the time domain (top) and frequency domain (bottom), and the columns to the right of it compare the results of Unrolled iPPG, UDEQ-iPPG, and DE-Prox-iPPG. The orange signal is the ground-truth pulse signal obtained from the finger, while the black signal is the input signal and the blue signals are the outputs of our algorithms. The frequency domain representation of the input signal in Figure~\ref{fig:comparison2} shows a noisy signal extracted from the face, and all three algorithms attenuate the noise. However, only UDEQ-iPPG and DE-Prox-iPPG recover the correct peak frequency coefficients with high signal-to-noise ratio. This provides evidence that our assumption about the frequency coefficients of the heartbeat signal---that its frequency coefficients are well represented as the fixed point of a denoiser---is a valid assumption.

\begin{figure*}
    \centering
    \includegraphics[width=0.80\textwidth]{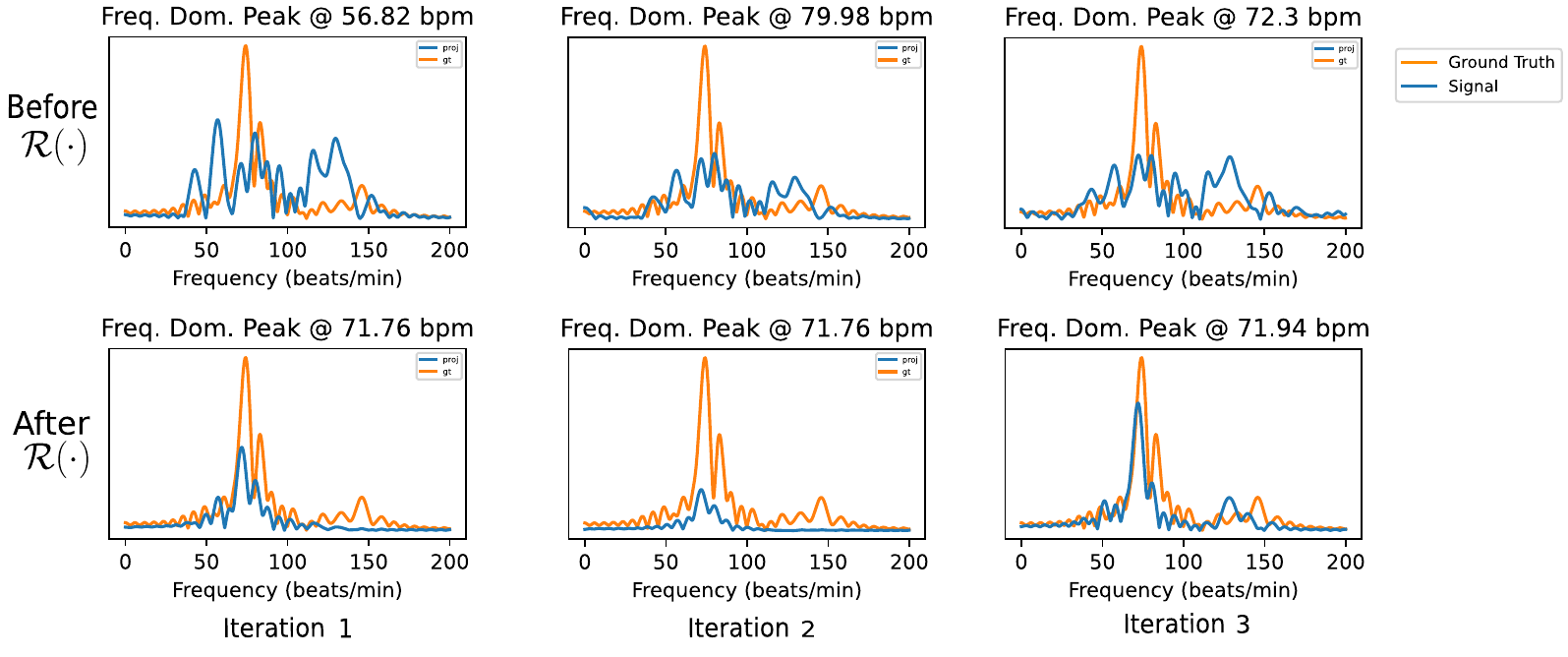}
    \caption{We show the spectrum of our recovered signal from facial video \textbf{(\textcolor{blue}{blue})} and the ground truth signal \textbf{(\textcolor{orange}{orange})} over three unrolling iterations on the PURE dataset using UDEQ-iPPG. For each iteration, the blue signal in the top row corresponds to $\mathbf{\Tilde{X}}_t$ before $\mathcal{R}(\cdot)$ acts upon it, while the blue signal in the bottom row corresponds to $\mathbf{X}_t$ after $\mathcal{R}(\cdot)$ as in   Figure~\ref{fig:unrolling_algorithm}. After multiple iterations, our algorithm captures the peak frequency and a harmonic while attenuating noise.}
    \label{fig:iterations_pure} 
\end{figure*}

During inference, our methods show rapid denoising and convergence to a fixed point, as shown in Figure~\ref{fig:iterations_pure}, where the orange signal is the frequency spectrum of the ground-truth signal, and the blue signal is the spectrum recovered by UDEQ-iPPG from facial video. We show the recovered signal over three unrolling iterations of UDEQ-iPPG. The blue signal in the top row shows $\mathbf{\Tilde{X}}_t$ from Figure~\ref{fig:unrolling_algorithm}, i.e., the signal after the gradient step but before the DEQ denoising step, while the the bottom row shows $\mathbf{X}_t$ after $\mathcal{R}(\cdot)$ operates upon it. Before the first iteration, the signal extracted from the face is noisy with an incorrect peak pulse-rate. In successive iterations, we notice the DEQ's ability to denoise the signal while the gradient step ensures consistency with the (noisy) extracted signal. By the end of three unrolling iterations, we successfully recover a cleaner signal. 

\begin{figure*}
    \centering
    \includegraphics[width=0.98\textwidth]{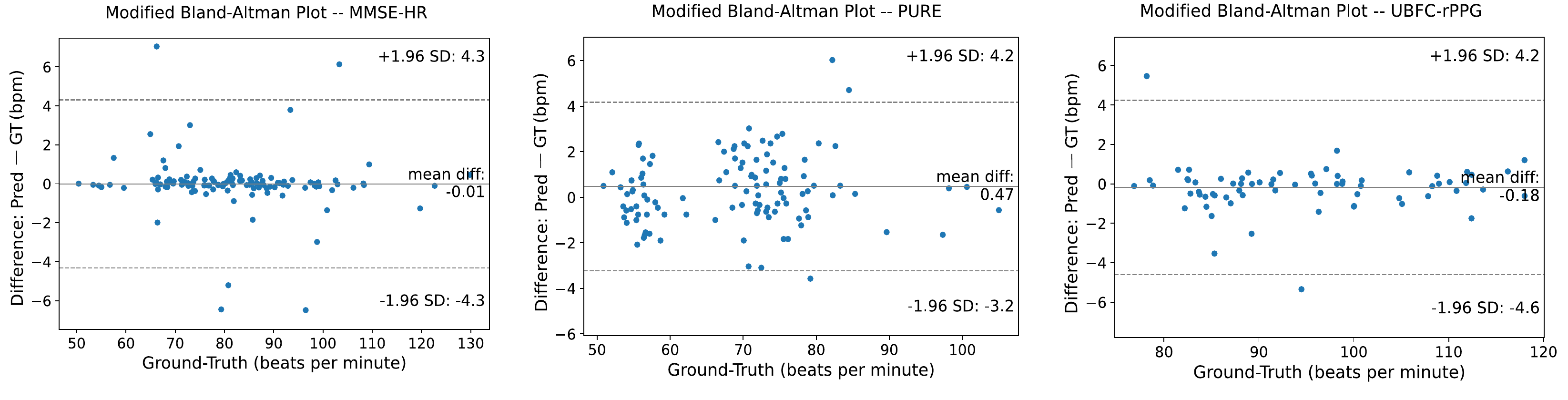}
    \caption{Bland-Altman Analysis between predicted and ground-truth pulse rate estimates for the MMSE-HR< PURE, and UBFC-rPPG datasets using the UDEQ-iPPG algorithm. Each time window is represented by one point, which plots the difference between the predicted and ground-truth heart rates vs. the ground-truth heart rate. Note that the 95\% limits of agreement, defined as the mean difference $\pm$ 1.96$\times$ the standard deviation, are within 5 bpm of the ground-truth heart rate on all datasets.}
    \label{fig:ba_mmse_pure}
\end{figure*}

\subsubsection{Statisical Analysis} To confirm the validity of our method, we perform a modified Bland-Altman analysis (also known as the Tukey Mean-Difference plot) between the estimated pulse rates and the ground-truth rate pulse rates for the UDEQ-iPPG method. In Figure~\ref{fig:ba_mmse_pure}, plot the difference between the predicted pulse rate and ground-truth pulse rate against the ground-truth pulse rate for the MMSE-HR, PURE, and UBFC-rPPG dataset. Each datapoint represents a single 30 s time window for the MMSE-HR dataset, a 10 s time window for the PURE dataset, and a 10 s time window for the UBFC-rPPG dataset. For all three datasets, our mean difference is close to zero, showing our method's agreement with the ground-truth. Furthermore, the 95\% limits of agreement, defined as the mean difference $\pm$ 1.96$\times$ the standard deviation, are less than or equal to 4.6 bpm from the ground-truth heart rate on all datasets. This shows strong performance, particularly over a range of pulse rates from 50 bpm to 130 bpm, confirming the effectiveness of our method.

\begin{table}[]
    \caption{Varying the type of model used for each denoiser, and the results on the three datasets.}
    \centering
    \setlength\tabcolsep{2pt}
    \resizebox{0.38\textwidth}{!}{
    \begin{tabular}{|c|c|c|c|ccc|}
    \hline
     Dataset & Method & $\mathcal{R}(\cdot)$ & $\mathcal{Q}(\cdot)$ & \textbf{MAE}  & \textbf{RMSE} & \textbf{PTE6}  \\
      & &  &  & bpm $\downarrow$ & bpm $\downarrow$& \% $\uparrow$ \\
    \hline
    \multirow{4}{*}{MMSE-HR} &
    --- & DEQ & DEQ &  1.21 & 3.02 & 93.02 \\
    
    & UDEQ-iPPG & DEQ & NN &  \textbf{0.95} & \textbf{2.53} & \textbf{94.57} \\
    
     & --- & NN & DEQ &  1.36 & 3.2 & 90.70 \\
    
    & Unrolled iPPG & NN & NN &  1.12 & 2.80 & 93.02 \\
    \hline
    \multirow{4}{*}{PURE} & ---
    & DEQ & DEQ &  0.30 & 0.80 & 100 \\
    
    & UDEQ-iPPG & DEQ & NN &  \textbf{0.18} & \textbf{0.35} & 100 \\
     
    & --- & NN & DEQ &  0.29 & 0.63 & 100 \\
    
    & Unrolled iPPG & NN & NN &  0.41 & 0.62 & 100 \\
    \hline
    \multirow{4}{*}{UBFC-rPPG} & ---
    & DEQ & DEQ &  7.10 & 10.10 & 63.64 \\
    
    & UDEQ-iPPG & DEQ & NN &  \textbf{0.46} & \textbf{0.82} & 100 \\
     
    & --- & NN & DEQ &  6.15 & 9.45 & 83.33 \\
    
    & Unrolled iPPG & NN & NN &  0.92 & 1.84 & 90.91 \\
    \hline
    \end{tabular}}
    \label{tab:deq_nn_comp}
\end{table}

\begin{table}[]
    \caption{Cross-dataset model evaluation}
    \centering
    \setlength\tabcolsep{2pt}
    \resizebox{0.48\textwidth}{!}{
    \begin{tabular}{|c|c|c|ccc|}
    \hline
    \textbf{Train} & \textbf{Test} & \textbf{Model} & \textbf{MAE (bpm)} $\downarrow$ & \textbf{RMSE (bpm)} $\downarrow$& \textbf{PTE6 (\%)} $\uparrow$ \\
    \hline
    \multirow{2}{*}{MMSE-HR} & \multirow{2}{*}{PURE} & Unrolled iPPG & \textbf{0.25} & \textbf{0.58} & \textbf{100} \\
    & & UDEQ-iPPG & 0.44 & 1.04 & 100\\
    \cline{3-6}
    \multirow{2}{*}{PURE} & \multirow{2}{*}{MMSE-HR} & Unrolled iPPG & 3.88 & 11.54 & 86.82 \\
    & & UDEQ-iPPG & \textbf{2.38} & \textbf{8.18} & \textbf{89.92}\\
    \hline
    \multirow{2}{*}{UBFC-rPPG} & \multirow{2}{*}{MMSE-HR} & Unrolled iPPG & 5.62 & 10.23 & 73.64 \\
    & & UDEQ-iPPG & \textbf{3.72} & \textbf{8.78} & \textbf{85.01}\\
    \cline{3-6}
    \multirow{2}{*}{MMSE-HR} & \multirow{2}{*}{UBFC-rPPG} & Unrolled iPPG & 2.32 & 8.46 & 90.69 \\
    & & UDEQ-iPPG & \textbf{0.42} & \textbf{1.26} & \textbf{100}\\
    
    \hline
    \end{tabular}}
    \label{tab:cross_dataset_results}
\end{table}

\subsection{Ablation Studies}\label{subsec:ablations}
\subsubsection{Model Class: DEQ versus Traditional Network}

First, we look at the effect of implementing learned denoising operator as DEQs and traditional neural networks in UDEQ-iPPG. We examine the four algorithmic configurations for model type where $\mathcal{R}(\cdot) \in \{\text{DEQ}, \text{NN}\}, \mathcal{Q}(\cdot) \in \{\text{DEQ}, \text{NN}\}$, where ``DEQ" is the fixed-point instantiation of the model in Figure~\ref{fig:architecture}, while ``NN" refers to a single application of the same model. The results are in Table~\ref{tab:deq_nn_comp}. The results show that the UDEQ-iPPG configuration, in which the signal $\mathbf{X}$ is modeled as the output of a fixed-point function ($\mathcal{R} =$ DEQ) and the noise $\mathbf{E}$ is modeled as unconstrained neural network ($\mathcal{Q} =$ NN), achieves state-of-the-art performance. The PURE dataset shows similar results; all model combinations perform well, but again the UDEQ-iPPG configuration performs best. On the UBFC-rPPG dataset, we see a clear distinction: when modeling the noise signal as a DEQ, our model struggles to converge. Both the Unrolled iPPG configuration ($\mathcal{R} =$ NN, $\mathcal{Q} =$ NN) and the UDEQ-iPPG configuration ($\mathcal{R} =$ DEQ, $\mathcal{Q} =$ NN) show good performance. Our intuition is that the pulse signal has structure, while the noise is unstructured; therefore, instantiating $\mathcal{R}(\cdot)$ as a DEQ and instantiating the noise $\mathcal{Q}$ as a traditional neural network learns best.

\subsubsection{Cross-Dataset Model Evaluation}Our methods' learned denoisers can generalize to unseen data; we show this in Table~\ref{tab:cross_dataset_results}, in which we train and test on different datasets. UDEQ-iPPG has an MAE of less than 4 bpm in all scenarios, while Unrolled iPPG does well in most scenarios.  We notice that UDEQ-iPPG transfers more effectively to other datasets, except in the case of MMSE-HR $\rightarrow$ PURE, when Unrolled iPPG performs better. Our algorithms transfer well to other datasets, though they suffer expected performance decreases when the domain shifts. We hypothesize that this happens because we learn \textit{the prior} associated with each dataset. Since this prior is data-driven, it may not capture the recording scenarios, noise, etc. of another dataset. This is clearly seen when training on datasets with less motion/illumination noise (i.e. PURE, UBFC-rPPG) and testing on datasets with more noise (MMSE-HR). 
\begin{table}[]
    \centering
    \caption{Mismatched gradient+denoiser iterations at inference for all models on the MMSE-HR dataset. Unrolled iPPG and UDEQ-iPPG are \textit{trained} on three unrolling iterations}
    \label{tab:iteration_modeling}
    \setlength\tabcolsep{2pt}
    \resizebox{.44\textwidth}{!}{
    \begin{tabular}{|c|c|ccc|}
    \hline
    \makecell{Test \\ iter} & \textbf{Method}  & \textbf{MAE (bpm)} $\downarrow$ & \textbf{RMSE (bpm)} $\downarrow$& \textbf{PTE6 (\%)} $\uparrow$ \\
    \hline
     \multirow{3}{*}{1} & Unrolled iPPG & 3.44 & 11.72 & 89.14 \\
     & UDEQ-iPPG &  2.72 & 10.20 & 90.69 \\
    & DE-Prox-iPPG & \textbf{1.28} & \textbf{2.86} & \textbf{93.80}\\
    \hline
    \multirow{3}{*}{2} & Unrolled iPPG & 3.18 & 11.11 & 89.14 \\
    & UDEQ-iPPG & 2.72 & 9.81 & 91.47\\
    & DE-Prox-iPPG & \textbf{1.22} & \textbf{2.80} & \textbf{93.80} \\
    \hline
    \multirow{3}{*}{3} & Unrolled iPPG & 1.11 & 2.97 & 93.53 \\
    & UDEQ-iPPG & \textbf{0.95} & \textbf{2.53} & \textbf{94.57} \\
    & DE-Prox-iPPG &  1.22 & 2.80 & 93.80\\
    \hline
    \end{tabular}}
\end{table}
\subsubsection{Mismatched Unrolling Iterations at Train and Test Time}
    One difference between UDEQ-iPPG and DE-Prox-iPPG is that UDEQ-iPPG is trained on a fixed number of \textit{unrolled} iterations, while DE-Prox-iPPG is run until solver convergence. In Table~\ref{tab:iteration_modeling}, we train Unrolled iPPG and UDEQ-iPPG on three unrolling iterations and train DE-Prox-iPPG until solver convergence, but limit the number of iterations \textit{at test time} for all models. We see that UDEQ-iPPG's performance deteriorates when training and testing on a different number of iterations. In contrast, DE-Prox-iPPG maintains its performance, even after just one iteration. However, we see rapid two-step convergence for DE-Prox-iPPG; this may indicate that the joint fixed-point of pulse and noise signals may reside in a minimum that is suboptimal for each signal independently but optimal for them together.

\begin{table*}[]
    \centering
    \caption{Mismatched unrolling iterations during training and testing for UDEQ-iPPG on the MMSE-HR dataset }
    \label{tab:unrolling_diff}
    \setlength\tabcolsep{2pt}
    \resizebox{.80\textwidth}{!}{
    \begin{tabular}{|c|ccc|ccc|ccc|ccc|}
    \hline
    & \multicolumn{12}{|c|}{Test Iterations}\\
    \hline
    \multirow{2}{*}{} & \multicolumn{3}{c|}{1 iter.} & \multicolumn{3}{c|}{3 iter.} & \multicolumn{3}{c|}{5 iter.} & \multicolumn{3}{c|}{10 iter.}\\
    \cline{2-13}
    \makecell{\textbf{Train} \\ \textbf{Iterations}} & \makecell{\textbf{MAE} \\ (bpm) $\downarrow$} & \makecell{\textbf{RMSE} \\ (bpm) $\downarrow$}&  \makecell{\textbf{PTE6}  \\ (\%)$\uparrow$} & \makecell{\textbf{MAE} \\ (bpm) $\downarrow$} & \makecell{\textbf{RMSE} \\ (bpm) $\downarrow$}&  \makecell{\textbf{PTE6}  \\ (\%)$\uparrow$} & \makecell{\textbf{MAE} \\ (bpm) $\downarrow$} & \makecell{\textbf{RMSE} \\ (bpm) $\downarrow$}&  \makecell{\textbf{PTE6}  \\ (\%)$\uparrow$} & \makecell{\textbf{MAE} \\ (bpm) $\downarrow$} & \makecell{\textbf{RMSE} \\ (bpm) $\downarrow$}&  \makecell{\textbf{PTE6}  \\ (\%)$\uparrow$} \\
    \hline
    1 & 1.51 & 3.65 & 90.69 & 2.72 & 8.58 & 89.92 &  2.12 & 7.17 & 90.69 & 2.08 & 7.11 & 90.79\\
    3 & 2.72 & 10.20 & 90.69 & \textbf{0.95} & \textbf{2.53} & \textbf{94.57} &  2.79 & 6.44 & 86.04 &  6.38 & 13.71 & 75.19\\
    5 & 2.61 & 9.81 & 90.69 & 1.03 & 2.65 & 94.57 & 1.28 & 3.28 & 93.02 &  6.35 & 15.30 & 75.19 \\
    10 & 2.78 & 10.00 & 89.14 & 1.01 & 3.53 & 96.12 & 1.66 & 4.80 & 90.69 &  1.10 & 3.08 & 95.34 \\
    \hline
    \end{tabular}}
\end{table*}

We explore this phenomenon further for UDEQ-iPPG.  In Table~\ref{tab:unrolling_diff}, we train UDEQ-iPPG using four different unrolling iterations; we then test using that same set of iterations for each model. The diagonal for the table shows the results of training and testing on the same number of unrolling iterations. We notice that training and testing on three unrolling iterations provides the best performance, and report those numbers in the Table~\ref{tab:allresults}. However, we notice that training on five and ten unrolling iterations and testing on three unrolling iterations provides better performance than training and testing with five unrolling iterations, or training and testing with ten unrolling iterations. We posit that this performance discrepancy happens due to overfitting---at test time, unrolling for more iterations overfits to the training data at the expense of generalization performance.

\subsubsection{Modeling the noise explicitly} Finally, to determine the effect of explicitly modeling the noise, we examine training with $\mathcal{R}(\cdot)$ only, i.e., solving the optimization problem
\begin{equation}\label{eq:opt_X_only}
    \begin{aligned}
        \min_{\mathbf{X}, \mathbf{E}} \frac{1}{2}\norm{\mathbf{Z - F^{-1}X}}_F^2   + \lambda \cdot \rho(\mathbf{X}).
    \end{aligned}
\end{equation}
We show the results in Table~\ref{tab:noise_modeling}. It is clear that Unrolled iPPG, DE-Prox-iPPG, and UDEQ-iPPG all benefit from modeling the noise in addition to the signal.

\begin{table}[]
    \caption{Heart rate estimation on the MMSE-HR dataset with and without modeling the noise component $\mathbf{E}$, where ``\xmark" solves Eq. \eqref{eq:opt_X_only} and ``\cmark" solves Eq. \eqref{eq:signal_recovery}.}
    \centering
    \setlength\tabcolsep{2pt}
    \resizebox{0.48\textwidth}{!}{
    \begin{tabular}{|c|c|ccc|}
    \hline
    \textbf{Method} & $\mathcal{Q}(\cdot)$? & \textbf{MAE (bpm)} $\downarrow$ & \textbf{RMSE (bpm)} $\downarrow$& \textbf{PTE6 (\%)} $\uparrow$ \\
    \hline
    \multirow{2}{*}{Unrolled iPPG} & \xmark &  1.44 & 3.87 & 92.25 \\
    & \cmark & \textbf{1.11} & \textbf{2.82} & \textbf{93.02} \\
    \hline
    \multirow{2}{*}{DE-Prox} & \xmark & 1.50 & 3.29 & 90.70\\
    & \cmark & \textbf{1.22} & \textbf{2.80} & \textbf{93.80} \\
    \hline
    \multirow{2}{*}{UDEQ-iPPG} & \xmark & 1.70 & 6.68 & 92.24 \\
    & \cmark & \textbf{0.95} & \textbf{2.53} & \textbf{94.57} \\
    \hline
    \end{tabular}}
    \label{tab:noise_modeling}
\end{table}

\section{Conclusion}
\label{sec:conclusion}

In this paper, we model pulse signal recovery from facial video as an inverse problem with learnable priors and propose solutions using unrolling methods, fixed-point methods, and a combination of unrolling and fixed-point methods. In our first method, we apply successive gradient descent and neural network-based denoising steps on the pulse signal and noise for a \textit{fixed number} of iterations---known commonly as ``unrolling" proximal gradient descent---and call our method Unrolled iPPG. Instead of applying a \textit{fixed} number of unrolling steps, Deep-Equilibrium-Proximal-iPPG (DE-Prox-iPPG), our second method, iterates both the gradient descent and denoising steps until fixed-point convergence. Finally, we combine unrolling and fixed-point methods in Unrolled-DEQ-iPPG, which applies the neural network-based denoising operator on the pulse signal until fixed-point convergence in each unrolling step. Our results show state-of-the-art performance using the UDEQ-iPPG,  while all three methods show good performance using significantly fewer parameters than previous methods. Future work should explore methods to determine the learnable prior associated with our learned denoising operators, adding a layer of interpretability to these state-of-the-art algorithms.

\bibliographystyle{IEEEtran}
\bibliography{main}

\end{document}